\documentclass[12pt,letterpaper]{article}

\usepackage[left=0.9in,right=0.9in,top=1in,bottom=1in]{geometry}
\usepackage[latin1]{inputenc}
\usepackage{amsmath}
\usepackage{amsbsy}
\usepackage{amsfonts}
\usepackage{amssymb}
\usepackage{graphicx}
\usepackage{multirow}
\usepackage{color}
\usepackage[linesnumbered,boxed]{algorithm2e}

\title{\bf Multi-Fingered Robotic Grasping: A Primer}
\author{Stefano Carpin\textsuperscript{1}, Shuo Liu\textsuperscript{1}, Joe Falco\textsuperscript{2}, Karl Van Wyk\textsuperscript{2}}
\date{\textsuperscript{1}University~of~California,~Merced\thanks{Stefano Carpin and Shuo Liu are partially supported by the National Institute of Standards and Technology (NIST) under 
Cooperative Agreement \# 70NANB12H142 ``Grasping and Simulation for Next-Generation Manufacturing Robots".
Any opinions, findings, and conclusions or recommendations expressed in these materials are those of
the authors and should not be interpreted as representing the official policies, either expressly or
implied, of the funding agencies of the U.S. Government.} \textsuperscript{2}National~Institute~of~Standards~and~Technology~(NIST)
\thanks{Certain commercial equipment, instruments,
or materials are identified in this article to foster understanding.
Such identification does not imply recommendation or endorsement
by NIST, nor does it imply that the materials or equipment
identified are necessarily the best available for the purpose}}

\begin{document}

\maketitle

\section{Introduction}

It has been argued that grasping, manipulation, and speech are among the most fundamental human
abilities unparalleled by animals \cite{BicchiKumarRamsete}. The importance of manipulation
and grasping is also evidenced by {\em ``the large fraction of the human motor cortex devoted
to manipulation and the number and sensitivity of mechanoreceptors in our palms
and fingertips"} \cite{OkamuraSurvey}.
Starting from these observations it is  unsurprising that grasping and manipulation
generated significant interest in the robotics community, and robotic grasping emerged as
an autonomous area of research since the development of the first 
robotic hands. Notwithstanding, robotic grasping continues to be one of the most
challenging areas in robotics. Its inherent difficulty is well captured by
Moravec's paradox, {\em ``It is comparatively easy
to make computers exhibit adult level performance on intelligence tests
or playing checkers, and difficult to give them the skills of a one-year old when
it comes to perception and dexterity"} \cite{Moravec1998}.\\

In robotics literature the terms {\em grasping} and {\em manipulation} are often 
used interchangeably, but
they indicate different challenges. Grasping deals with the problem of restraining an object with a robotic hand or gripper.
One of the foremost problems in grasping is deciding where the robot end-effector should make contact with the 
object to be grasped and which forces should be applied in order to ensure certain properties, like the ability to resist external 
disturbances.

Manipulation is instead concerned with the problem of determining how a body can be moved by a robot.
The object to be moved may be firmly held by the robot (hence the connection to grasping), but this does
not necessarily need to be the case. For example, a robot may manipulate an object by pushing.
Manipulation problems can then be addressed using algorithms and models coming from the motion 
planning domain \cite{LatombeMotionPlanning,ChosetBook,LaValleBook}. There is no
universally accepted definition for the expression {\em dexterous manipulation}. However,
many authors use the term to indicate manipulation problems where an object is 
moved using robotic fingers, either with one or two hands \cite{OkamuraSurvey,DollarICAR2011}.\\

Given the continued advancements in the field, the complexity of the problem, and its practical importance, literature in robotic 
grasping is vast and features some very detailed textbooks \cite{MasonBook,MurrayLiSastryManipulation} and
surveys \cite{PrattichizzoHandbook,BicchiKumar,OkamuraSurvey,ShimogaSurvey,BicchiTransactions2000}.
In this primer we focus on multi-fingered hands and we are particularly attentive to performance metrics and
test methods. Moreover, we are interested in methods and algorithms that can be applied 
in the real world, as opposed to abstract formulations that are difficult
to use in practical applications. One should notice upfront that the divide 
between theory and application is still significant in certain areas of grasping.

\section{Grasp modeling}
\label{sec:formalism}
As in virtually every  area of robotics, it is necessary to develop and introduce an appropriate 
mathematical formalism to study the grasping problem. It shall be noted that the level of
sophistication achieved in this area is remarkable and a complete discussion is beyond the scope 
of this primer. The interested reader is for example referred to \cite{MurrayLiSastryManipulation} for an
in-depth discussion. We therefore limit the number of concepts we introduce to the minimum and we
assume the reader is familiar with basic concepts from classic mechanics.
Our goal is not to offer a complete treatise, but rather to introduce the minimum amount of
notation needed to formalize concepts and algorithms presented later on.

\subsection{Kinematics}
\label{sec:kinematics}

\begin{figure}[htb]
\centering
\includegraphics[width=5cm]{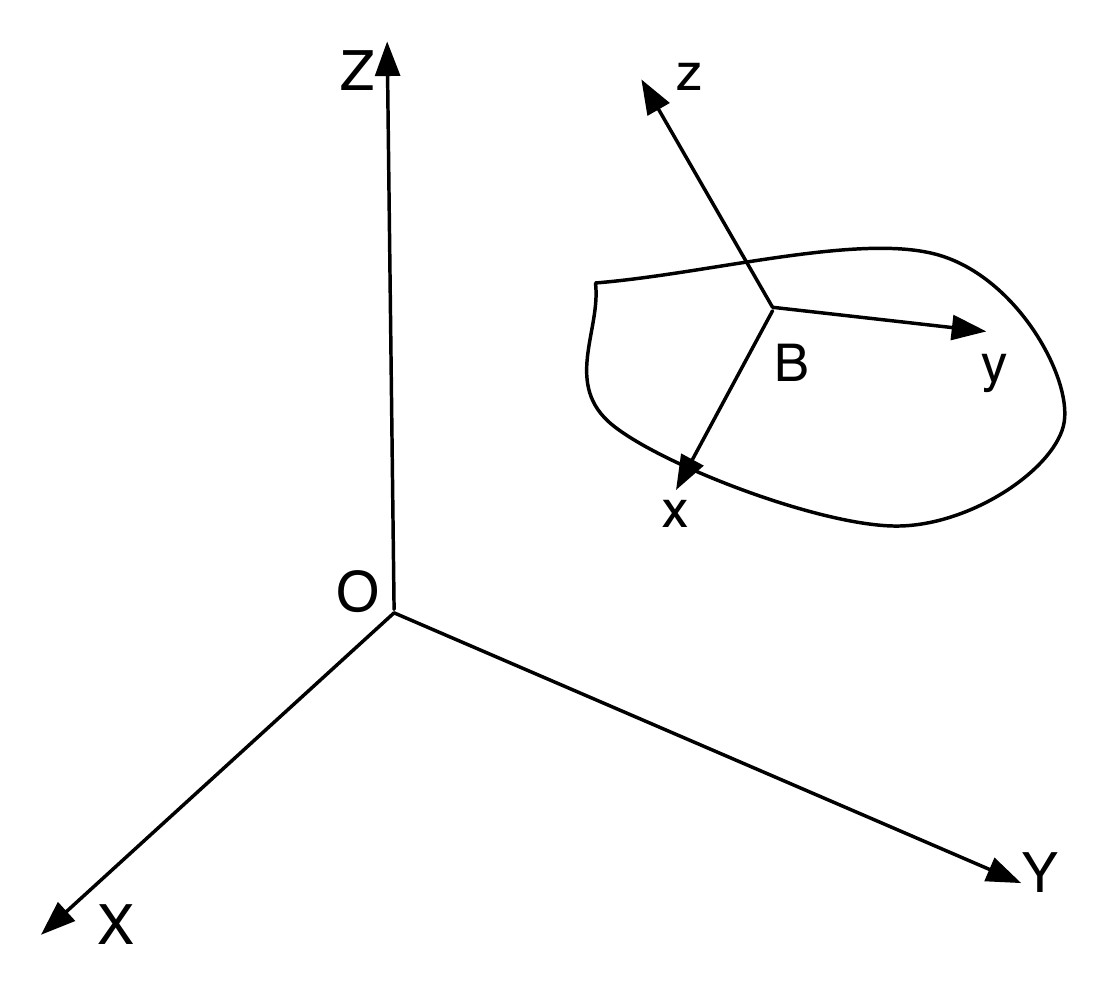}
\caption{World frame and body frame}
\label{fig:bodyframe}
\end{figure}

Throughout this primer we assume the object to be grasped is  a rigid body $\mathcal{B}$.
Assuming an inertial reference frame $O_{XYZ}$ is given, we need to specify the pose (i.e., position
and orientation) of $\mathcal{B}$ with respect to $O_{XYZ}$. The classic approach is to 
rigidly attach a frame to $\mathcal{B}$ (commonly called
{\em body frame} or {\em moving frame}) and to then describe the pose of this body-fixed frame.
In principle, this body-fixed frame could be attached anywhere to $\mathcal{B}$, but in the grasping literature, it is customary to assume this frame is placed at the center of mass of $\mathcal{B}$.
Let $B_{xyz}$ be the body frame attached to $\mathcal{B}$ (see Fig. \ref{fig:bodyframe}). 

Three parameters are used to describe the position of the body frame and
three parameters are needed for its orientation. Therefore, the pose of $\mathcal{B}$ is
fully described by a 6-dimensional vector. Different representations can be used for 
the orientation (see for example \cite{CraigBook,LaValleBook} for  in depth discussions about different 
parametrizations for the space of rotations). Generically, the 6-dimensional pose vector\footnote{Throughout this document we assume vectors are column vectors.} for $\mathcal{B}$ is defined as $\mathbf{u}$ (so-called {\em operational space} \cite{OrioloBook}),

\[
\mathbf{u} = 
\left[
\begin{array}{c}
\mathbf{p}\\
\mathbf{\Phi}
\end{array}
\right]
\]

\noindent where $\mathbf{p}=[p_x~p_y~p_z]^T$ represents the translational position and $\mathbf{\Phi}$ represents the orientation. The orientation can be expressed through Euler angles\footnote{One could in fact also use a non minimal representation, e.g., quaternions to describe
the orientation. In this case $\mathbf{q}$ would then have 7 dimensions. This option will not be explored.}.
 The vector $\mathbf{u}$ is a more compact  representation
of the  homogeneous transformation matrix

\[
_B^OT =
\left[
\begin{array}{cccc}
& & & p_x\\
& _B^OR & & p_y\\
& & & p_z\\
0 & 0 & 0 & 1
\end{array}
\right]
\]
where $_B^OR$ is the rotation matrix describing $\mathcal{B}$'s orientation with respect to $O_{XYZ}$
and $(p_x,p_y,p_x)$ define the position of the origin of $B_{xyz}$ with respect to $O_{XYZ}$.
It is important to note that one can define suitable mappings between 
$\bf u$ and $_B^OT$. Therefore depending on the context, either of these forms will be used.
\\

\begin{figure}[htb]
\centering
\includegraphics[width=0.4\linewidth]{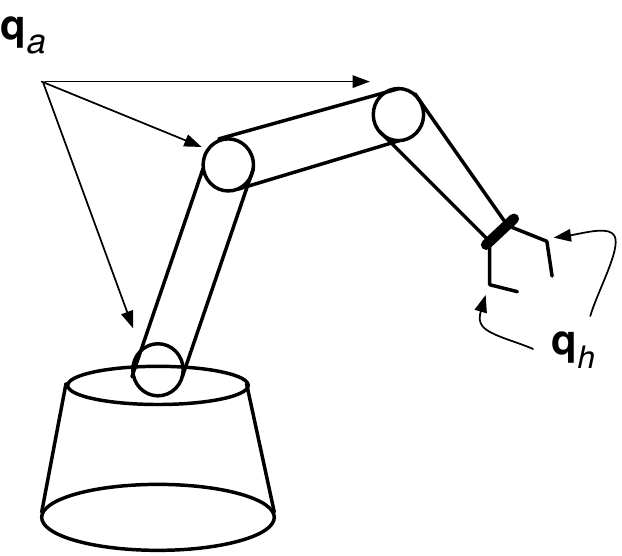}
\caption{Vector $\mathbf{q}$ includes both the degrees of freedom defining the arm pose $\mathbf{q}_a$ and the hand configuration 
 $\mathbf{q}_h$. }
\label{fig:RobotWithHand}
\end{figure}

To study grasping problems a compact representation for the pose of the robotic hand $\mathcal{H}$ is needed, too. The 
most convenient model uses a vector of $m$ joint displacements to fully specify the hand placement in space.
Note that $m$ is the overall number of degrees of freedom, which includes both the degrees of freedom
of the arm and of the hand.  
It is customary to indicate this $m$-dimensional vector as $\mathbf{q}$ and to
indicate its components as $q_i$, i.e., $\mathbf{q} = [q_1,q_2,\dots,q_m]^T$. 
The vector $\mathbf{q}$ is said to belong to the {\em joint space}.
In certain situations it will be necessary to consider just the degrees of freedom of the hand.
Therefore it  is convenient to assume $m=a+h$, where $a$ is the number of degrees of freedom
of the arm and $h$ is the number of degrees of freedom of the hand. The vector $\mathbf{q}$
can then be decomposed as $\mathbf{q} =[\mathbf{q}_a^T ~\mathbf{q}_h^T]^T$, where $\mathbf{q}_a$ includes
just the degrees of freedom of the arm and $\mathbf{q}_h$ includes the degrees of freedom of the hand (see 
Fig. \ref{fig:RobotWithHand}).

Given $\mathbf{q}$, forward kinematics can be used to determine the position of any point on the robot arm or hand 
(for example using the Denavit-Hartenberg method \cite{CraigBook}). 
In grasping applications, the pose of some points are particularly relevant. These are the palm of the
robotic hand, and the tips of the fingers. Once fixed frames are assigned to the these points, their
pose can also be determined through forward kinematics. The hand orientation, in particular, is often 
assumed to be the orientation of the frame attached to the palm. This information is important when parametrizing grasps.\\

\begin{figure}[htb]
\centering
\includegraphics[width=6cm]{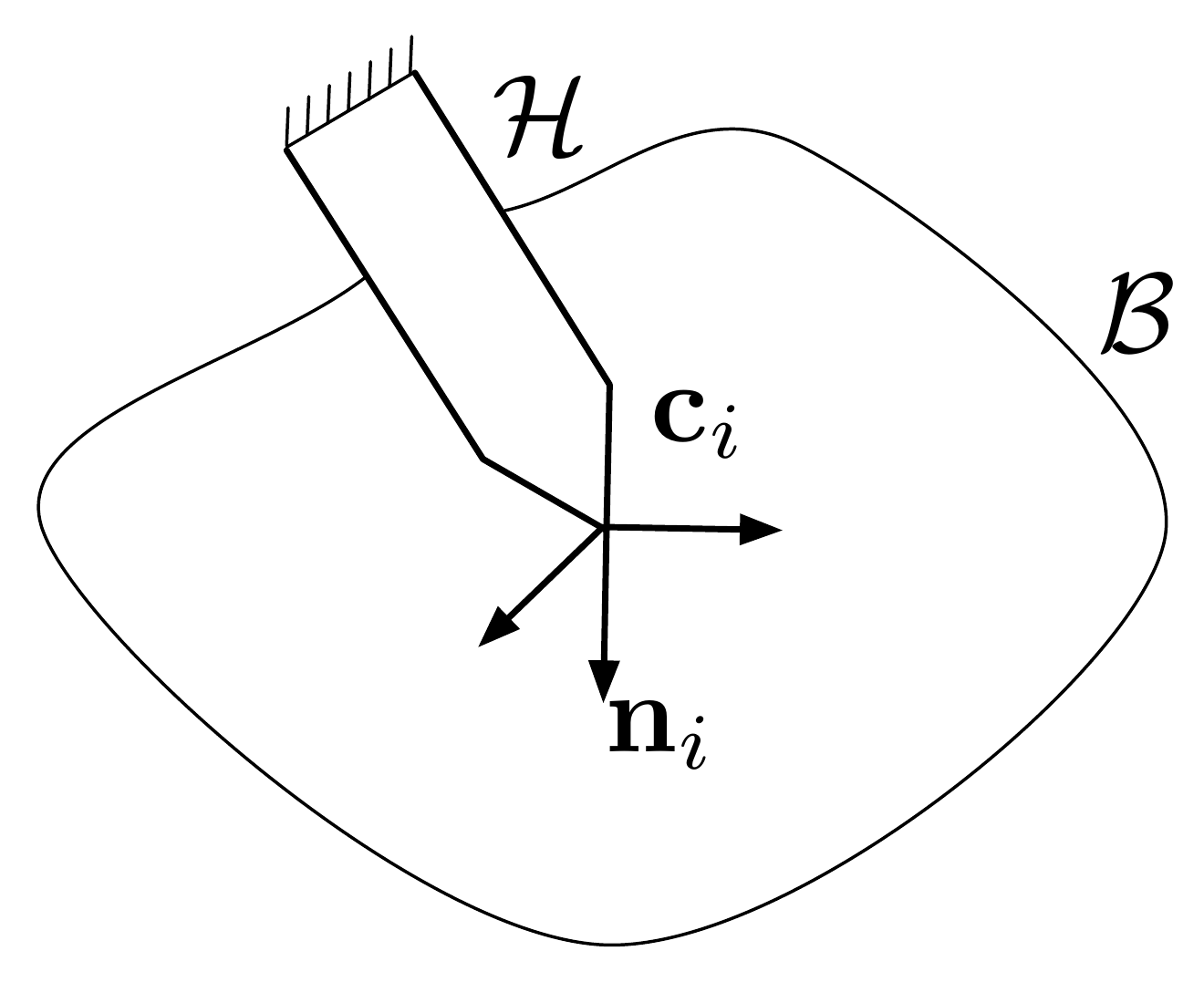}
\caption{The $i$-th finger of hand $\mathcal{H}$ makes contact with $\mathcal{B}$ at point $\mathbf{c}_i$. At the contact point 
a frame is defined with the axis $\mathbf{n}_i$ normal to the plane tangent to $\mathcal{B}$ at point $\mathbf{c}_i$. The other two axes
are on the tangent plane.}
\label{fig:contact}
\end{figure}

Grasping relies on contacts between the hand $\mathcal{H}$ and the body $\mathcal{B}$. Although in general
a contact between $\mathcal{H}$ and  $\mathcal{B}$ may occur in arbitrarily shaped regions, it is 
customary to only model situations where the contact occurs at a single point. More complex scenarios,
like for example contact along a segment, can be dealt with by using an appropriate set of contact points.
The $i$-th contact point is indicated with the letter $\mathbf{c}_i$ and its three dimensional coordinates 
can be expressed with respect to $O_{XYZ}$ or $B_{xyz}$. 
In the following, different models to characterize friction at a contact point will be
presented. To this end, it is  convenient to assign frames to contact points, too. We assume that 
for every contact point $\mathbf{c}_i$, the tangent plane to $\mathcal{B}$  is defined. The frame assigned
to $\mathbf{c}_i$ has one axis pointing inside $\mathcal{B}$ along the normal to the tangent plane, 
and the other two axis lying on the tangent plane. The normal axis will be indicated as $\mathbf{n}_i$ (see Figure \ref{fig:contact}).
Let $n_c$ be the number of contact points.
The role of contact points is critical as a conduit for transmitting forces and moments on $\mathcal{B}$ by the robotic hand.

\subsection{Motion and forces}
In grasping applications it is common to  rely on 
{\em screw theory} to describe the motion of rigid objects. Three concepts will be introduced, namely, {\em screw}, {\em twist}, and
{\em wrench}. 
Screws, twists, and wrenches can be introduced through differential geometry
and a wealth of results can therefore be used to characterize them.
The book by Murray et al. \cite{MurrayLiSastryManipulation} offers a very thorough 
analysis of these concepts based on exponential representations.  However, in the following
we stick to a more classic approach. It is important to note that while these concepts
have been widely used, definitions and representations are not always consistent.\\

A result known since 1830 and normally referred to as Chasles's theorem states that  any spatial displacement of 
a rigid body can be decomposed as a rotation about an axis followed by a translation along the same axis.
A displacement could then be described with three parameters, namely an axis $\mathbf{a}$, an angle $\theta$, and a displacement
$d$. However, it is common not to provide $d$, but to rather specify a pitch $h$, defined as the ratio
between $d$ and $\theta$. Once the angle $\theta$ is given, the corresponding displacement can be computed as $\theta h$.
Collectively, $\mathbf{a}$, $h$, and $\theta$ define a {\em screw} (also called {\em screw motion}).
At this point it should also be clear why the name {\em screw} is used to describe this geometric object\footnote{In  \cite{MasonBook}
screws, twists, and related objects are introduced using slightly different definitions. While eventually one can reconcile these
differing representations, we here stick to the definitions given in \cite{MurrayLiSastryManipulation} and \cite{PrattichizzoHandbook}.}.
Note that a screw with infinite pitch corresponds to a pure translation, whereas a screw with 0 pitch
is a pure rotation. \\

\noindent 	A {\em twist} is the representation of the velocity of a body $\mathcal{B}$ and is a six dimensional vector 
\[
\mathbf{t} = 
\left[
\begin{array}{c}
\mathbf{v}\\
{\boldsymbol \omega}
\end{array}
\right]
\]
where $\bf v$ is the translational velocity and $\boldsymbol \omega$ is the rotational velocity. 
Both $\bf v$ and  $\boldsymbol \omega$ are vectors with three components.
\\

Finally, a {\em wrench} is used to model a force $\bf f$ applied to $\bf p$ on the surface of $\mathcal{B}$. The point $\bf f$ may be a contact point in case the force is applied by one
of the fingers, but also a different point in case the force is due to an external load.
 The wrench is also
a six dimensional vector including both the force $\bf f$ and the moment $\bf m$ generated by $\bf f$.
 
 \[
 \mathbf{w} = 
 \left[
 \begin{array}{c}
 \mathbf{f}\\
 \bf m
 \end{array}
 \right].
 \]
Note that both twists and wrenches can be described in different reference frames and not only with respect to $O_{XYZ}$.
In particular, they can be expressed in $B_{xyz}$ or a  frame associated with a contact point.
Moreover, since $\bf w$ includes a moment component $\bf m$, when defining a wrench it is
necessary to specify the point about which $\bf m$ is defined.
 The concept of a
wrench is fundamental to formalize the notion of a force-closed grasp, and to study
how forces are exchanged between the robotic hand and the object through the contact points.  If a multi-fingered robotic hand $\mathcal{H}$
makes contact with $\mathcal{B}$ at multiple points, there will be multiple sources of interaction forces, each with its own associated wrench. The mapping between forces and wrenches
is linear, and if all wrenches are expressed with respect to the same frame, the wrenches can be superimposed by addition. Wrench models will be discussed once appropriate friction models are described in the next section.

\subsection{Friction}

Contact points are usually characterized with three different types of friction \cite{BicchiKumar,BurdickHandbook}. 
It should be noted up front that this taxonomy is a useful approximation, but does not model all possible contacts
and is defined only for rigid body objects.

\begin{itemize}
\item {\em Frictionless} contact point. In this case the finger can only exert a force along ${\bf n}_i$,
the normal to the surface at the contact point. In particular, the force is directed towards the
inside of the object, i.e., the finger can push but not pull. Therefore, frictionless contact
points introduce a constraint on the force of the type $f_n \geq 0$, where $f_n$ is the 
force along the normal.

\item {\em Frictional} contact point. In this case, the finger can  exert a force along
 the normal to the surface but also along the tangent plane. 
Different friction  models could be used, but the most commonly used  is based on Coulomb's law.
Coulomb's law states that no relative motion between the finger and the object happens  as long as
the tangential component of the force $f_t$ satisfies the following inequality
\begin{equation}
f_t \leq \mu f_n
\label{eq:coulomb}
\end{equation} 
where $f_n$ is the normal component and $\mu$ is the static friction coefficient characterizing the
 coupling between the finger and the object. According to this model, Eq. \eqref{eq:coulomb} defines
 the largest tangential force that can be applied. Since this force can be arbitrarily oriented
in the tangent plane, a common way to graphically depict this situation is given by the so-called
{\em friction cone}. The friction cone has its apex at the contact point and is oriented along the
normal to the tangent plane. Its aperture  is $\tan^{-1} \mu$ (see Figure \ref{fig:FrictionCone}).
Often, the tangential component $f_t$ is expressed in terms of its 
two projections $f_1$ and $f_2$ along the axis defined
on the tangent plane. In this case we can then write $f_t = \sqrt{f_1^2+f_2^2}$ and Eq. \ref{eq:coulomb}
can then be rewritten as $ \sqrt{f_1^2+f_2^2} \leq \mu f_n$. Moreover, the inequality $f_n \geq 0$ 
must still hold. These conditions define the friction cone, $FC$, at contact point ${\bf p}_i$ (see also Fig. \ref{fig:FrictionCone}):
\[
\textrm{FC}({\bf p}_i) = \left\{ [f_1~f_2~f_n]^T \in \mathbb{R}^3 ~|~\sqrt{f_1^2+f_2^2} \leq \mu f_n \wedge f_n \geq 0\right\}
\]

\begin{figure}[htb]
\centering
\includegraphics[width=5cm]{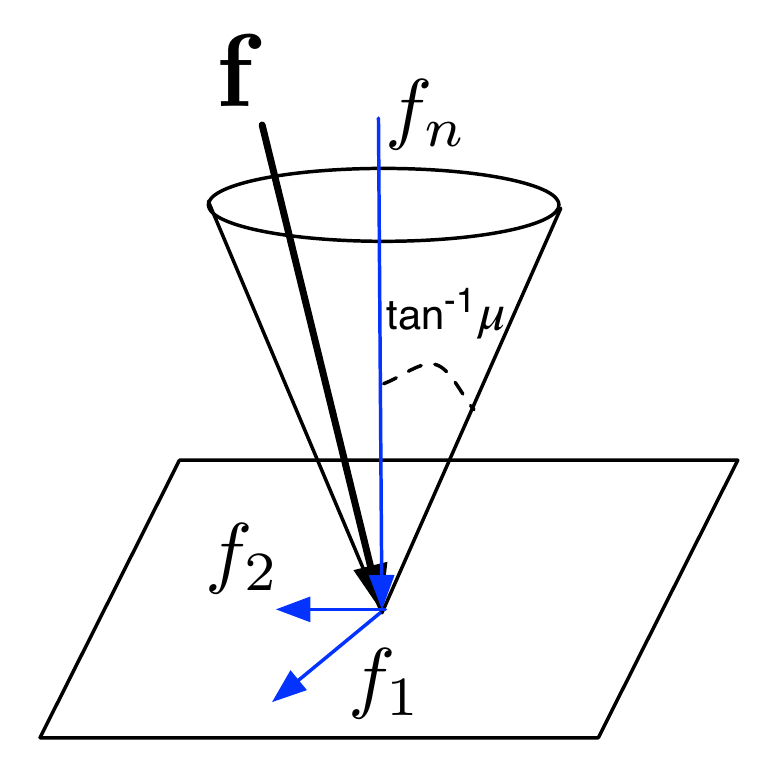}
\caption{Friction cone}
\label{fig:FrictionCone}
\end{figure}

Sliding between the finger and the object should not occur as long as the total force, $F$, exerted by finger is 
inside the cone. Note that in many practical situations the cone is approximated with 
a pyramid \cite{FerraryCanny,ZhuTransactions2003}.

\item {\em Soft} contact. In this case the finger can not only exert normal and tangential
forces, but also a torsional moment about the normal to the contact point. Note that in this case
we usually talk about a contact area rather than a contact point. For a soft contact the same
inequalities specified for a frictional contact point must be satisfied. Moreover, indicating
with $\tau$ the intensity of the torsional moment, then $\tau \leq \gamma f_n$.
\end{itemize}

Although there is agreement about the provided taxonomy, nomenclature is not standard.
For example, frictionless contact points are also called {\em point contact without friction},
whereas frictional contact points are also referred to as {\em hard finger} \cite{PrattichizzoHandbook}.\\

The type of contact determines then the number of independent wrenches that can be exerted on the
body $\mathcal{B}$ through the contact. For a frictionless contact point, just one wrench
can be exerted, and this is the result of the force applied
along the normal to the tangent on the contact point.
For a frictional contact point, three independent wrenches can be applied. One of these
is again due to the force along the normal to the tangent plane, whereas the other two are 
due to the forces exerted on  
the tangent plane. Finally, for the soft contact, four independent wrenches can be applied. These
are due to three forces (same as the frictional point), and one torsional moment about the
 normal to the contact
point. When the contact is more complex, the number of independent wrenches can still
be determined, though this is more complicated \cite{SalisburyPhDThesis}.\\

\begin{figure}
\centering
\includegraphics[width=6cm]{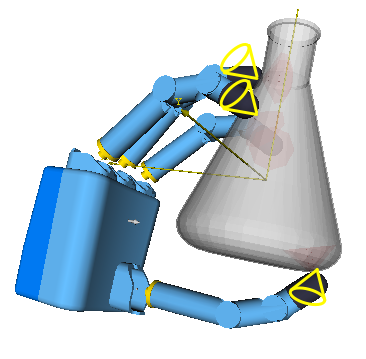}
\caption{Multifingered grasp with frictional contact points. The same friction model is assumed at each contact point
and therefore friction cones are associated with each finger-object contact.}
\label{fig:frictioncones}
\end{figure}

In robotic grasping it is customary to assume that the same friction model applies to each of the contact points (e.g., Figure \ref{fig:frictioncones}), but this is not necessarily the case.

\subsection{Multi-fingered grasps}
\label{sec:multifingered}
Given a friction model, a multi-fingered grasp can be formally specified using 
a representation to determine the type of restraint 
obtained by the grasp. When there are $n_c$ contact points
between the fingers and the body, just specifying their coordinates ${\bf c}_1,\dots,{\bf c}_{n_c}$
is not sufficient, because it is also necessary to consider the forces exchanged
through the contacts and how the fingers can be moved to exert these forces.

This information is encoded in two matrices, namely the hand Jacobian ${\bf J}$ and the
grasp map ${\bf G}$. The hand Jacobian ${\bf J}$ is the classic Jacobian
matrix studied in robot kinematics. Let ${\bf q}_h$ be the $h$ degrees of freedom of the hand.
Then the position of the fingertips can be determined thorough 
forward kinematics. Forward kinematics is then a function $\textrm{FK}: \mathbb{R}^h \rightarrow \mathbb{R}^{6n_c}$
mapping the $h$ degrees of freedom ${\bf q}_h$ into the frames attached to the fingertips\footnote{In
this case we assumed that the frame attached to each fingertip is represented using the minimal
representation $\bf u$ in operational space.}. 
Note that we assumed 
there are $n_c$ fingers, to use the same symbol utilized for the number of contact points.
The hand Jacobian J can be mapped from joint velocities q into frame velocities using
the relationship ${\bf J}{\dot{ \bf q}}$.

While $\bf J$ is used to express relationships involving velocities, the grasp matrix $\bf G$
defines how forces exerted at the contact points are mapped into wrenches.  $\bf G$
 is specified using a friction model that includes the forces and moments that are applied to the object.

The relationship between friction models and wrenches is then discussed in the following.
\begin{enumerate}
\item Consider a frictionless contact point ${\bf c}_1$ and let  ${\bf f}_1$ be the force
applied. Because the point is frictionless, ${\bf f}_1$ can only be applied along the normal
to the tangent plane at ${\bf c}_1$ and the wrench is
\[
{\bf w}_1 = \left[
\begin{array}{c}
{\bf f}_1\\
 ({\bf c}_1-{\bf o}) \times {\bf f}_1)
\end{array}
\right]
\]
where $\bf o$ is the point about which moments are computed. This is typically the origin of $B_{xyz}$.
In section \ref{sec:kinematics} we assumed that a frame is associated to every
contact point with an axis $\bf n$ normal to the tangent point defined at the contact point.
Then, recalling this frame we can rewrite ${\bf w}_1$ as follows, where we evidence the dependence
on the force intensity $f_1$.
\[
{\bf w}_1 = \left[
\begin{array}{c}
0\\
0\\
1\\
-p_y\\
p_x\\
0
\end{array}
\right]
f_1
\]
where $p_x$ and $p_y$ are the coordinates of ${\bf c}_1$ with respect to frame $B_{xyz}$.
In this second form, the wrench matrix is written as the product of a {\em wrench basis} with 
a scalar.
\item Next, consider a frictional contact point ${\bf c}_2$. Three forces can be exerted, one 
along the normal and the other two on the tangent plane. 
Let these forces be ${\bf f}_1$,${\bf f}_2$ and ${\bf f}_3$, where the first two are on 
the tangent plane and the third is the normal force. These three forces are not independent
from each other but, rather, are related by the friction coefficient, as 
implied by the  Coulomb inequality (Eq. \ref{eq:coulomb}).
Stated differently, their sum must lie inside the friction cone FC.
 Applying a reasoning similar 
to the one we used for the frictionless contact point, we can then write an expression for the
wrench:
\[
{\bf w}_2 = \left[
\begin{array}{ccc}
{\bf f}_1 & {\bf f}_2& {\bf f}_3\\
({\bf c}_1-{\bf o}) \times {\bf f}_1 & ({\bf c}_2-{\bf o}) \times {\bf f}_2   & ({\bf c}_3-{\bf o}) \times {\bf f}_3 
\end{array}
\right].
\]
This expression can also be rewritten as the product between a wrench basis and a vector
with  the three intensities $f_1,f_2$ and $f_3$:
\[
{\bf w}_2 = \left[
\begin{array}{ccc}
1 & 0& 0\\
0 & 1& 0\\
0 & 0& 1\\
0 & p_z & -p_y\\
-p_z & 0  & p_x\\
p_y & -p_x & 0
\end{array}
\right]
\left[
\begin{array}{c}
f_1\\
f_2\\
f_3
\end{array}
\right]
\]
\item Finally, consider a soft contact point ${\bf c}_3$. In this case the wrench not only includes
three forces as for the frictional point, but also one torsional moment. Let $f_1,f_2$ and $f_3$ be the
intensities of the forces and $m$ the intensity of the moment. The wrench and
the wrench basis are:
\[
{\bf w}_3 = 
\left[
\begin{array}{cccc}
{\bf f}_1 & {\bf f}_2& {\bf f}_3 & 0\\
 ({\bf c}_1-{\bf o}) \times {\bf f}_1  & ({\bf c}_2-{\bf o}) \times {\bf f}_2   &  ({\bf c}_3-{\bf o}) \times {\bf f}_3  & {\bf m}
\end{array}
\right]
\]
\[
=
\left[
\begin{array}{cccc}
1 & 0& 0 & 0\\
0 & 1& 0 & 0\\
0 & 0& 1 & 0\\
0 & p_z & -p_y & 0\\
-p_z & 0  & p_x & 0 \\
p_y & -p_x & 0 & 1
\end{array}
\right]
\left[
\begin{array}{c}
f_1\\
f_2\\
f_3\\
\tau
\end{array}
\right]
\]
\end{enumerate}

Given a set of $n_c$ contact points, each defines its own wrench matrix and basis. 
The grasp matrix ${\bf G}$ 
is the matrix obtained by composing the various wrench bases side by side. It is therefore 
a matrix with six rows and the number of columns is $n=n_l + 3n_f + 4n_s$, where $n_l$ is the
number of frictionless contact points, $n_f$ is the number of frictional contact points, and
$n_s$ is the number of soft contact points. The grasp matrix $\bf G$ can be seen
as a linear map $G:\mathbb{R}^n \rightarrow \mathbb{R}^6$ mapping the intensity of forces and moments
into the overall wrench acting on $\mathcal{B}$. To exploit this matrix
notation, it is then convenient to use an $n$-dimensional vector ${\bf l}$ including all 
the force and moment components. In this case, the resulting wrench can then be written as
$\bf Gl$, where we will implicitly assume that $\bf l$ satisfies the constraints imposed
by the generalized friction code FC.\\

Equipped with this notation, a multifingered grasp $\mathcal{G}$ 
is then fully specified by ${\bf J}$, $\bf G$,
and the friction coefficients needed to define the friction cone FC. Note that $\bf G$ implicitly provides the contact points ${\bf c}_i$, so these do not need to be specified separately.

\subsection{Form and force closure}
The concepts of form and force closure 
emerged as the two main characterizations for a grasp.
Both ideas find their origin in mechanism design and predate robotics and grasping.
In fact, they are more than 100 years old (see \cite{Reuleaux} for a translation
of the original book). However,
grasping problems have stimulated research aiming to produce accurate and efficient tests
to determine if a grasp achieves any of these two properties. 
Before dwelling upon the details, it is useful to anticipate 
 that every form closure grasp is also a force closure grasp, but not viceversa.

\subsubsection{Form closure}
Informally speaking, form closure is obtained when the robotic hand surrounds the object to be grasped 
in such a way that the object cannot move without colliding with the hand. 
A different way to express the same concept is saying that contacts prevent all possible motions of $\mathcal{B}$,
including infinitesimal motions \cite{TrinkleICRA1992}.
This idea can be formalized 
in different ways. For example, one could say that form closure occurs when the object is located at a singular 
configuration in its configuration space \cite{MasonBook}. Alternatively, form closure can be analytically defined 
as follows \cite{PrattichizzoHandbook}. Let $\mathbf{u}$ be the configuration of the object and $\mathbf{q}$
be configuration of the hand. Assuming there are $n_c$ contact points between the hand and the object, a 
gap function $\psi_i(\mathbf{u},\mathbf{q})$, $1\leq i \leq n_c$ can be defined for each of the contact points.
The gap function gives the distance between the hand and the object at the contact point. By definition,
 $\psi_i(\mathbf{u},\mathbf{q})=0$ when the contact occurs, becomes positive  when the contact breaks
 (the finger moves away from the object), and becomes negative when there is penetration (the finger enters
 the object). Based on this definition, a grasp achieves form closure if and only if the following implication
  holds:
\begin{equation}
\psi( \mathbf{u}+d\mathbf{u},\mathbf{q}) \geq 0 \Rightarrow d\mathbf{u}=0
\label{eq:form_closure}
\end{equation}
where inequalities between vectors have to be interpreted component-wise.
It is easy to see that this equation is equivalent to say that the object cannot move ($d\mathbf{u}=0$), i.e., that it is located
at an isolated point in its configuration space.  Note that this definition does not rely on friction, i.e., form closure is
achievable irrespectively of friction. This class of grasps is therefore also indicated as {\em frictionless grasps}.\\

\noindent {\bf Form closure tests.} A practical problem of great importance is knowing whether a certain 
set of contacts achieves  form closure or not. A related problem is ranking various grasps leading to
force closure from the best to the worst, according to some performance metric, like for example resistance to 
external disturbances. Numerous criteria have been developed for the case where planar objects are being grasped
(see e.g., \cite{VanDerStappenIJRR2000}). In this paper we however concentrate exclusively on the three dimensional
scenario.

A result formulated by Somov more than 100 years old states that at least seven contact points 
are necessary to obtain form closure 
with an arbitrarily shaped object with six degrees of freedom, i.e., for the 
general problem of objects moving in three dimensions.
Because this is a necessary condition, this simple criterion provides an easy test to quickly determine
that all grasps relying on six or less points cannot obtain form closure on a generic object. 
But the condition is not sufficient,
and grasps with seven contact points can still be insufficient to achieve form closure. More sophisticated 
tests are therefore necessary.
However, if one restricts the class of objects being grasped, then different results hold. 
For example, seven contacts points are necessary and sufficient for
the special case of a three dimensional polyhedron \cite{Papadimitriou1990}.\\

Form closure tests of different order can be developed and 
are introduced  as follows.
Consider the first order Taylor
expansion of Eq. \ref{eq:form_closure} around $(\mathbf{u},\mathbf{q})$. This can be written
as (higher order terms have been omitted for brevity)

\[
\psi( \mathbf{u}+d\mathbf{u},\mathbf{q}) =
 \psi( \mathbf{u},\mathbf{q}) + \left. \frac{\partial \psi({\bf u},{\bf q})}{\partial {\bf u}} \right|_{({\bf u},{\bf q})}d{\bf u} = 
  \left. \frac{\partial \psi({\bf u},{\bf q})}{\partial {\bf u}} \right|_{({\bf u},{\bf q})}d{\bf u}
\]
where we used the fact that by definition $\psi( \mathbf{u},\mathbf{q})=0$ at a contact configuration. This expansion can then be
related to Eq. \ref{eq:form_closure} as follows.  If there exist $d {\bf u}$ such that 
$\frac{\partial \psi({\bf u},{\bf q})}{\partial {\bf u}}d{\bf u}$ has at least one positive component, then
the grasp does not achieve form closure. In fact, in this case $d {\bf u}$ identifies a direction along which
the object can be moved causing one of the components of the gap functions to become positive, hence breaking one of the contacts.

Alternatively,  one can consider the negated equivalent\footnote{That is, rather than using
$p \Rightarrow q$ we consider the logically equivalent implication $\neg q \Rightarrow \neg p$}  of the implication given in Eq. \ref{eq:form_closure}. This means
that  for every choice of $d{\bf u}$ different from 0, then $\frac{\partial \psi({\bf u},{\bf q})}{\partial {\bf u}}d{\bf u}$
has at least one negative component. In this case the grasp then achieves form closure.\\

When neither  of these two conditions hold, the first order Taylor expansion is insufficient to classify the grasp.
In this case a Taylor expansion of  order two should be considered. If using the second order Taylor expansion,
neither of the two conditions hold, then a third order Taylor expansion should be used, and so on.
 For this reason,
in grasping literature it is common to read about form closure tests of order $n$, with reference to the considered 
order for the Taylor expansion. Rimon and Burdick provide numerous results discussing the importance of
higher order analysis in restraint analysis \cite{RimonBurdickTRO98,RimonBurdickTRO98Bis,RimonICRA1996}.\\

The first order form closure test  leads to a linear programming program whose solution not only determines
whether a set of contact points achieve form closure, but also provides a quantitative measure that can be
used to rank different grasps.
This idea was first introduced by Trinkle in \cite{TrinkleICRA1992,TrinkleTransactions1992}
and the reader is referred to those papers for details.

\subsubsection{Force closure}
Intuitively, a grasp achieves force closure if it can be maintained in spite of external forces acting on the
restrained object. This idea can be rigorously formalized, but it should be immediately noticed that
there exist slightly different definitions for this concept.\\

A concise definition of force closure can be given using the grasp matrix $\bf G$ \cite{MurrayLiSastryManipulation}. A grasp achieves force closure if it can balance
any external wrench ${\bf w}_e$ applied to the object. Algebraically, this means that
for each external wrench ${\bf w}_e$, there exist a vector ${\bf l} \in \textrm{FC}$ such that
\begin{equation}\label{eq:force_closure_condition}
{\bf Gl} = -{\bf w}_e
\end{equation}

\begin{figure}
\centering
\includegraphics[width=5cm]{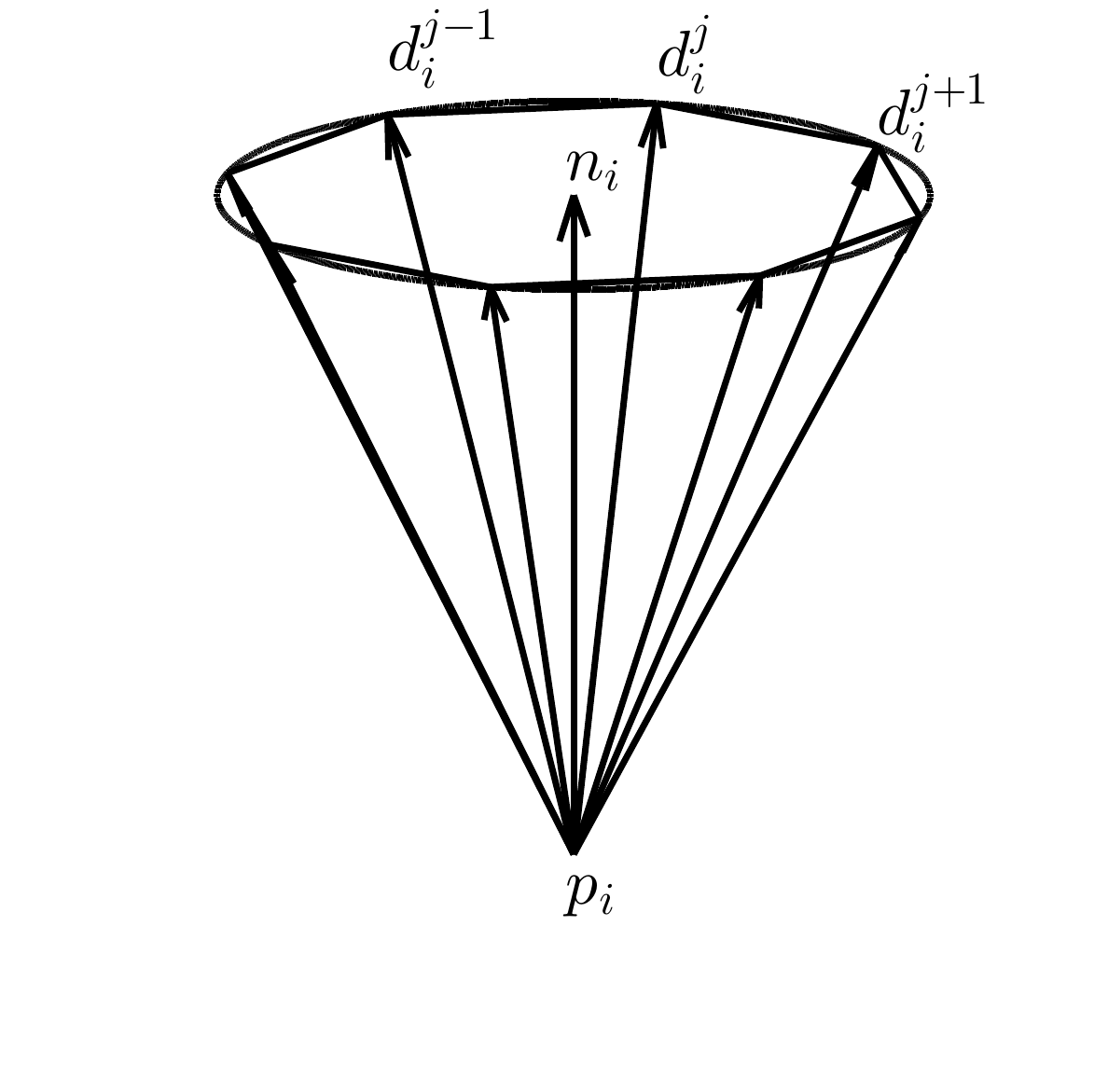}
\caption{Friction cones are commonly approximated with regular pyramids}
\label{fig:cone_approx}
\end{figure}

From this algebraic definition it is possible to immediately derive algebraic conditions
to determine if a given grasp achieves force closure. For example, a necessary condition
is that ${\textrm{rank}}({\bf G})=6$. However this condition is just necessary but not 
sufficient, because ${\bf l}$ is restricted to be in the friction cone FC, and this 
is a subset of $\mathbb{R}^n$. A necessary and sufficient condition is that ${\bf G}( \textrm{FC})=\mathbb{R}^6$,
where ${\bf G}( \textrm{FC})$ is the set obtained by applying the grasp map to every point in the friction cone
FC. Another useful characterization used in practice relates the force closure to convex hulls, which is particularly useful when introducing grasp quality metrics.
As formerly stated, the friction cone is often approximated with a regular pyramid (see Fig. \ref{fig:cone_approx}).

Using this approximation, each contact
force lying within  the friction cone can be written as 
a  non-negative combination of forces along the boundary of the friction cone, i.e., 
$\mathbf{f}_i = \sum_{j=1}^{k} \alpha_{i,j}\mathbf{f}_{i,j}$ with $\alpha_{i,j} \geq 0$.
 Based on this discretization the wrench generated by the $i$th contact
force can then be written as
\[
\mathbf{w}_i=\left[
\begin{array}{c}
\mathbf{f}_i\\
(\mathbf{c}_i - \mathbf{o}) \times \mathbf{f}_i
\end{array}
\right] = 
\left[
\begin{array}{c}
\sum_{j=1}^{k} \alpha_{i,j}\mathbf{f}_{i,j}\\
(\mathbf{c}_i - \mathbf{o}) \times \sum_{j=1}^{k} \alpha_{i,j}\mathbf{f}_{i,j}
\end{array}
\right]=
\]
\[
= \sum_{j=1}^{k} \alpha_{i,j}
\left[
\begin{array}{c}
\mathbf{f}_{i,j}\\
(\mathbf{c}_i - \mathbf{o}) \times \mathbf{f}_{i,j}
\end{array}
\right] = \sum_{j=1}^{k}\alpha_{i,j}  \mathbf{w}_{i,j}
\]
where each of the $\mathbf{w}_{i,j}$ is called {\em elementary wrench}. The $kn_c$ elementary wrenches
$\mathbf{w}_{i,j}$ can be arranged in a $6\times kn_c$ matrix $\mathcal{G}$ that approximates the {\em grasp matrix}. 
As per Eq. \eqref{eq:force_closure_condition}, the grasp achieves force closure if  each disturbance wrench $\mathbf{w}_e$ 
can be countered by forces inside the friction cones.
That is to say,  the grasp matrix $\mathcal{G}$
positively spans $\mathbb{R}^6$. In this case, it is known that an equivalent condition for force closure 
 is that the convex hull of all the $kn_c$ elementary wrenches $\mathbf{w}_{i,j}$ includes the origin
 (see e.g., \cite{MurrayLiSastryManipulation}).

\subsubsection{Grasp quality}
From the definitions of form and force closure, it is evident that these 
conditions could be satisfied by multiple different grasps.
The availability of multiple solutions to the same problem  immediately 
 raises the issue of  solution selection. In particular, it 
is natural to look for a {\em grasp quality metric} to rank different 
solutions and then pick the best one. This concept has been heavily 
investigated for grasps achieving force closure, and in the following 
discussion we limit our discussion to this case.  The reader is referred to
\cite{GraspAuro2014} for a comprehensive review of grasp metrics. \\

The most commonly used criterion is the so-called Ferrari-Canny
metric \cite{FerraryCanny}. For sake of correctness it should be noticed that similar ideas were already explored in \cite{Kirkpatrick}.
The underlying idea is that given two grasps and an external load, 
one should prefer the grasp that can resist the load by exerting the 
smallest effort. In general, since the external load is not known, the criterion should be applied in a worst-case manner, i.e., 
by picking the grasp that can be resist {\em any} external load with 
the smallest effort. The concept of effort needs to be formalized as well.
In \cite{FerraryCanny} two different definitions are put forward.

\noindent Any wrench exerted on $\mathcal{B}$  through the $n_c$ contacts can be scaled to 
belong to the set
\begin{equation}\label{eq:wlinf}
W_{L_{\infty}}=\textrm{CH}\left(\bigoplus_{i=1}^{n_c} \{\mathbf{w}_{i,1}\dots \mathbf{w}_{i,k} \}\right)
\end{equation}
where $\oplus$ represents the Minkovski sum and CH stands for {\em Convex Hull}. From an operative point of view
it is worth observing that the inner Minkovski sum in Eq. \ref{eq:wlinf} gives
a set of finite elements, and therefore $W_{L_{\infty}}$ is the convex hull of
a finite set of elements in $\mathbb{R}^6$.
The grasp metric $Q_{\infty}$ is then defined
as the distance of the closest facet of $W_{L_{\infty}}$ from the origin, i.e., 
\begin{equation}\label{eq:qinfydefinition}
Q_{\infty} = \inf_{\mathbf{x}\in \partial W_{L_{\infty}}} ||\mathbf{x}||_2
\end{equation}
where $\partial W_{L_{\infty}}$ indicates the boundary of $W_{L_{\infty}}$, i.e., the union of its facets.
Physically, 
$Q_{\infty}$ aims to minimize the maximum force exerted by any of the $m$ fingers
to resist an arbitrary external wrench.
Alternatively, one could aim at minimizing the sum of forces exerted by all fingers.
To this end, consider
\begin{equation}\label{eq:wl1}
W_{L_1}=\textrm{CH}\left(\bigcup_{i=1}^{n_c} \{\mathbf{w}_{i,1}\dots \mathbf{w}_{i,k} \}\right).
\end{equation}
Similarly to $Q_{\infty}$, the grasp metric  $Q_{1}$ is then defined
as the distance of the closest facet of $W_{L_1}$ from the origin
\begin{equation}\label{eq:q1definition}
Q_{1} = \inf_{\mathbf{x}\in \partial W_{L_{1}}} ||\mathbf{x}||_2.
\end{equation}

Both these ideas can be easily formalized considering the convex hull
of the friction cones at the contact points. 
Moreover, their computation is also related to the computation of the convex
hull.
In particular, if the
origin of $\mathbb{R}^6$ is not inside the convex hull, the grasp does
not achieve force closure. However, when the origin is inside, the
distance between the origin and the closest face of convex hull 
is chosen as the grasp quality, and grasps with larger values are
preferred. Indeed, grasps with larger values of this metric can resist
external loads using wrenches with smaller magnitude.  

Various relevant aspects should be mentioned. First, practically speaking, the effort definition used is defined as the magnitude of the largest wrench generated by any finger.
Next, to increase the computational efficiency,
 the friction cone is discretized using a representation based
on a regular pyramid. A tradeoff between accuracy and speed evidently
emerges, with more edges on the pyramid leading to a more accurate result,
but also requiring more computational time. As discussed later on, in many
instances the value of the grasp quality metric is used to drive the 
grasp planning process, and the quality is then iteratively computed
multiple times. To this end, significant efforts have been devoted to 
accelerate the computation of the quality metric. The QuickHull
algorithm  \cite{QuickHullArticle} had emerged as the de-facto standard in this
area thanks to its efficiency and its widespread availability. To compute the
metric, one starts computing the convex hull in six dimensions and then
computes the distance between the origin and the hull. However, it
is easy to realize that this approach is wasteful inasmuch as it computes
the full convex hull. In reality, just the face closest to the origin is
needed to compute the metric. Capitalizing on this observation, 
we recently introduced the Partial Quick Hull algorithm \cite{Liu2}
that greatly accelerates the computation of these metrics.

Despite its popularity, the Ferrari-Canny metric suffers from various drawbacks.
For example, it is not scale invariant since it depends on the choice of the point about
which torques are computed and it does not take into account the geometry of the
object. Moreover, it is in general too conservative, i.e., it scores a grasp based on
its ability to resist an arbitrary disturbance wrench, whereas in practice the set
of realistic disturbance wrenches is much smaller than the set of all disturbance
wrenches. Starting from these observations, Strandberg and Wahlberg proposed
an alternative metric based on the ability of a grasp to resist wrenches occurring
in practice \cite{strandberg2006}. Their method builds upon the concept of
object wrench space defined by Pollard \cite{Pollard1994}. The object wrench space
is the set of wrenches that can be exerted on an object through the action of a
disturbance force. Evidently this set is different from the set of all wrenches, since
for example is does not contain wrenches with just a torque component but no force
component. Therefore, in \cite{strandberg2006}, it is proposed to score a grasp on
the ability to resist these wrenches. This revised metric overcomes most of the limitations
encountered by the Ferrari-Canny metric, but in practice it has been rarely used
because its computation is demanding. In fact, Borst in \cite{borst2006}
proposed a method to approximate its value, but thse alternate
algorithms did not receive much attention either. Recently, Liu and Carpin \cite{LiuCASE2015}
proposed an exact algorithm to compute the metric proposed in \cite{strandberg2006}.
The algorithm is based on the principle of partial quick hull computation
and builds upon some of the ideas developed in \cite{Liu2}.

\subsection{Grasp parametrizations}
\label{sec:grasp_parametrizations}
A growing number of grasping methods rely on the assumption that a set of grasps is 
computed \textit{a priori} and can be queried at run time for retrieval (see Section \ref{sec:algorithms}).
To implement this functionality it is therefore necessary to chose an appropriate representation
of a grasp to simplify storage and retrieval. Note that this cannot just be ${\bf q}$ or ${\bf q}_a$
because it is necessary to somehow also consider the spatial relationship between the hand
and the object to be grasped without necessarily storing the object itself.
An ideal parametrization should be as general as possible and not be tied to a specific robot hand.
Also, one should notice that although these representations are known as {\em grasp} parametrizations,
they should not be confused with the formerly introduced concept of multi-fingered grasp (see 
Section \ref{sec:multifingered}). 
In comparison, they describe the pose of the hand as it approaches an object before performing the actual grasp.\\

We present two different parametrizations proposed in grasping literature.
Berenson and colleagues \cite{BerensonHumanoids2007} propose to parametrize each
grasp with four parameters and assume a body frame has been attached to palm of the robot hand.
Although the specifics of this frame are not explicitly spelled out, it is convenient to assume that 
one of the axes of this frame, say ${\bf n}_p$, is orthogonal to the palm\footnote{It is then
implicitly assumed that it is possible to identify the direction orthogonal to the palm. This may be
easy for certain robotic hands, e.g., the Barret Hand, but less obvious for other devices, so this choice 
is not unambiguous. }. The four parameters are the following:
\begin{itemize}
\item the direction of approach of the hand, i.e., ${\bf n}_p$;
\item a point $\bf p$ on the surface of $\mathcal{B}$ towards which the hand is moving;
\item the roll $\theta$ of the hand about ${\bf n}_p$;
\item the preshape of the hand, i.e., ${\bf q}_h$.
\end{itemize}
Note that ${\bf n}_p$ and $\theta$ can be computed from ${\bf q}_a$ through forward kinematics, but also observe
that ${\bf q}_a$ includes additional information that is not necessary for this parametrization.\\

Another parametrization is proposed in \cite{AllenICRA2004} by Pelossof and colleagues. Although specifically
designed for the Barrett hand, it can be generalized and extended for other hands, too.
The Barrett arm and hand has $m=10$ degrees of freedom, with $a=6$ and $h=4$. Note that for the Barrett Hand
three of the four values in ${\bf q}_h$ define the opening of the fingers, whereas the last parameter 
gives the spread.
In order to reduce the number of parameters, the authors make some assumptions. The first is
that the palm should be parallel to the surface of the object, i.e., ${\bf n}_p$ should
be orthogonal to the surface. Moreover, because the hand will move along ${\bf n}_p$, the distance
between the object and the hand does not need to be explicitly considered.
It is also assumed that the hand starts with its fingers fully 
open, i.e., three of the four values in ${\bf q}_h$ are fixed upfront. Based on these assumptions,
four parameters can be used to parametrize a grasp:
\begin{itemize}
\item two values, say $\eta$ and $\zeta$, to specify the position of the palm of the hand. Note that 
because it is  assumed  that  ${\bf n}_p$ is  orthogonal to the surface (i.e., it is constrained
to be on a plane)  and the distance to the object is not relevant, two parameters suffice.
\item the roll $\theta$ of the hand about ${\bf n}_p$;
\item the spread of the fingers, i.e., the only free parameter in ${\bf q}_h$.
\end{itemize}

A few comments about these two alternatives are needed. The first remark is about the size of the
parametrization itself. While they both nominally rely on four parameters, it is clear that the dimensionality
of the first one is much higher, because both  ${\bf n}_p$ and $\bf p$ each require three parameters to be specified.
In addition, the first parametrization offers the possibility to fully specify the preshape with ${\bf q}_h$ and
this comes at the cost of an increase in its dimension. Both representations include the roll angle 
about ${\bf n}_p$. In \cite{BerensonHumanoids2007} the authors maintain that their parametrization is 
superior to the one proposed in \cite{AllenICRA2004}. In particular they argue that it is advantageous 
to relax the constraint that ${\bf n}_p$ is orthogonal to the surface. This claim is not controversial
because indeed it allows to explore a wider range of possibilities. 
Furthermore, the parametrization proposed in \cite{BerensonHumanoids2007} offers more
flexibility in specifying the preshape. It is clear however that these benefits come at the cost of increasing
the dimension of the parametrization and potentially of the solution search space, and therefore depending
on the grasp planning algorithm used, the cost may overcome the benefits.

\section{Grasp Planning Algorithms}
\label{sec:algorithms}
Grasp planning algorithms have been investigated since the very introduction of robotic grippers. 
Through the years more and more complex problems and situations have been considered.
 In terms 
of robot actuators, initial studies mostly dealt with parallel jaw grippers, but nowadays a significant amount of
research is devoted to multi-fingered hands with three or more fingers and numerous degrees of freedom.
The complexity of objects considered in grasp problems has also evolved and  the focus is now on
grasping objects used in everyday activities by humans, as opposed to early investigations
concerned with planar objects with elementary shapes. At the same time, algorithms considering different degrees of 
uncertainty (e.g., uncertainty in the location or shape of the object to be grasped, or imprecise robot motions, etc.) 
have been developed. Coherently with the goals of this primer, we will not consider the 
problem of grasping a 2D object\footnote{Technically speaking, 2D objects do not exist. This expression 
is however commonly used to indicate flat objects in which one of the dimensions is much smaller than the others.}. 
In terms of hands, our focus is on multi-fingered hands, but we 
will nevertheless consider certain algorithms that have been demonstrated on parallel-jaw grippers
if their applicability to multi-fingered hands is evident. \newline

Research in grasping and existing methods could be categorized in many different ways. We developed
a taxonomy presented in Table \ref{tab:taxonomy} and we  structure this 
section accordingly\footnote{References
given in Table \ref{tab:taxonomy} are meant to be just a sampling of the vast literature in the field. 
They provide starting points to delve into this area.}. 
This is just one way of characterizing work in this area, and given the variety of algorithms
 we are aware that some will not clearly fit into it. Nevertheless,
for the class of problems relevant for this effort, we found that this table offers a useful
way to look at how the grasping problem can be approached.\\

\begin{table}[htb]
\centering
\begin{tabular}{l|l|l|l|l|l|l}
\hline
\multicolumn{7}{|c|}{{\bf Grasping algorithms}} \\ \hline

\multicolumn{3}{|p{6cm}|}{{\bf Model Based}\newline The geometry or shape of the object to be grasped is given a-priori. Planning happens without integrating sensor information.} & \multicolumn{4}{|p{8cm}|}{{\bf Model-less}\newline Shape is not known a-priori and the robot uses sensors to acquire information about the object to be grasped.} \\ \hline

\multicolumn{2}{|p{4cm}|}{{\bf Geometric only}} & \multicolumn{1}{|p{2cm}|}{{\bf Data Driven}} & \multicolumn{2}{|p{4cm}|}{{\bf Vision based}} & \multicolumn{2}{|p{4cm}|}{{\bf Touch based}}\\ \hline

\multicolumn{1}{|p{2cm}|}{{\bf Consider Object Only} \newline \cite{AllenICRA2007}\cite{LiSastryJournalRA}\cite{AllenICRA2003}  \cite{AllenICRA2004}\cite{PonceTransactions1995}\cite{ZhuTransactions2003}} &  \multicolumn{1}{|p{2cm}|}{{\bf Consider Object and Environment}\newline \cite{BerensonHumanoids2007}} & \cite{ColumbiaGraspDatabase,AllenDataDrivenGrasping}& \multicolumn{1}{|p{2cm}|}{{\bf Construct model} \newline \cite{collet:2009}\cite{DuneIROS2008}\cite{GallardoICAR} \cite{HubnerICRA2008}\cite{ChinellatoAutRob2008} } &\multicolumn{1}{|p{2cm}|}{{\bf Operate in feature space}\newline \cite{KragicIJRR2012}\cite{SaxenaNgIJRR}}&   \multicolumn{1}{|p{2cm}|}{{\bf Construct model} \newline  \cite{BierbaumTactile2008}\cite{BierbaumTactile}\cite{GrupenTRO2010}}& \multicolumn{1}{|p{2cm}|}{{\bf Operate in feature space } \cite{AllenLearningGraspStability}\cite{AllenICRA2011}} \\ \hline
\end{tabular}

\caption{A possible taxonomy for grasping algorithms with references to methods falling in each category.
Note that not all referenced papers are fully covered in this primer.}
\label{tab:taxonomy}
\end{table}

A final word on the use of machine learning algorithms is warranted. In general, the use of machine learning 
in robotics (including robotic grasping) is increasing at a fast pace. 
We believe three reasons 
justify this expansion. First, dramatic improvements in computational power and major algorithmic 
breakthroughs have produced a wide variety of techniques that can now efficiently solve
hard computational problems. Second, the necessity of developing robots operating in
poorly structured environments or subject to various sources of noise calls for the use
of algorithms and representations inherently capable of dealing with probabilistic uncertainty. 
In this area, too, machine learning offers a wide variety of tools. Finally, the ever increasing
availability of inexpensive sensors producing vast amounts of data enables the development 
of machine learning algorithms that extract models from the data. 
The interplay between machine learning and
robotic grasping appears to be unavoidable in the future and is evidenced by the selected papers
we discuss in the following. The increasing interplay between machine learning and robotics
is well captured by the emerging expression {\em cloud robotics}, whereby the cloud is
used to enable learning algorithms operating over large, remotely stored data sets \cite{GoldbergTASE2015}.

\subsection{Model-based Algorithms}

In this subsection we discuss grasp planning methods sharing the common assumption that a geometric
model for the object to be grasped is provided in advance. Many different approaches
exploiting this knowledge have been explored. At one end of the spectrum, we find methods relying 
 on geometric models only. Typically, these are models for the hand, the object to be grasped, and the
environment. We name this class {\em geometric} algorithms. At the other end of the spectrum, we find algorithms that exploit also additional knowledge, 
like past experience acquired while grasping, pre-computed grasps, or data collected while executing
the grasp plan itself. We dub this second class {\em data driven} methods, and they will
be discussed in the second part.\\

The overarching computational question is to determine an appropriate set of contact points
on the surface of the object being grasped in order to achieve a certain objective, like e.g.,
form closure or force closure. Many parameters need to be given to fully specify an instance
of this problem, like the shape of the object (e.g., polyhedral or with curvature), friction
model, number of fingers of the hand, and so on. Therefore, many different algorithms have
been proposed in the robotics literature. Before dwelling upon the details, it is worthwhile
recalling that we are here interested in planning (or synthesis) algorithms, as opposed
to analysis algorithms. The former aim to produce a set of contact points satisfying certain
constraints, whereas the latter are intended to determine if a given set of points satisfy the
constraints or not.

\subsubsection{Geometric algorithms for grasp point calculation}
Some common features are shared by algorithms that plan grasping points for an arbitrary
number of fingers on three dimensional objects. A common pattern among these
approaches is to formulate the problem as a search of the grasp configuration case
with the intent of optimizing a certain objective function. According to this paradigm,
then, grasp point calculation is the result of an optimization problem. Despite the
apparent simplicity of this paradigm, a certain level of mathematical sophistication
is required to formulate meaningful objective functions.\\

Li and Sastry \cite{LiSastryJournalRA} propose a method to compute a set of 
grasp points that aims at
maximizing some grasp quality indexes discussed or introduced in the paper. 
Their method works for 3D objects
but has the limitation of being applicable only for objects with a polyhedral boundary.
The approach relies on the screw theory extensively discussed in \cite{MurrayLiSastryManipulation}.
Three different quality indexes are considered. The first two are similar to other
measures discussed in  literature, whereas the third one is novel and developed by 
the authors. The first one, indicated as $\delta$, is defined as the smallest
singular value of the grasp matrix $\bf G$. The second one, indicated as $\nu$, is
the so-called volume in wrench space and is defined by the following integral
\[
\nu = \int_{G(B^n_1\cap \textrm{FC})} d\nu,
\]
where $B^n_1$ is the unit ball in $\mathbb{R}^n$, $ \textrm{FC}$ is the friction cone, and
$G(B^n_1\cap \textrm{FC})$ is the image of the set through the map defined by the grasping
matrix $\bf G$. Both $\delta$ and $\nu$ are known to have desirable properties in terms of 
grasp performance measurement, but Li and Sastry outline drawbacks that render them inappropriate
to drive the grasp-point calculation process. In particular, $\delta$ is not invariant
under a change of the torque origin, and $\nu$ does not capture force closure (which the
authors call stability). In other words, $\nu$ in itself does not discriminate between
grasps achieving closure, but can be used to rank grasps only after they have been
shown to obtain closure. To overcome these limitations, the authors introduce
a third quality measure that is {\em task oriented}. This term indicates that the measure
aims at identifying the ability of a grasp to generate wrenches relevant to the task
being performed. To be precise, tasks are modeled by ellipsoids in the wrench space.
The intuition is that this ellipsoid should capture the type of wrenches necessary
to successfully complete a given task, and to make their case the authors
compare the wrenches that should be exerted to complete a {\em peg-in-hole} task
with the wrenches to perform some writing with a pencil. Evidently, these two tasks
require different wrenches and then should be associated with different ellipsoids.
A task ellipsoid is defined starting from a vector ${\bf a} \in \mathbb{R}^6$ and
a $6\times 6$ positive definite matrix $\bf Q$:
\[
A_{\beta} = \{ {\bf w }\in \mathbb{R}^6, {\bf w}^T{\bf Qw} + {\bf w}^T{\bf a} \leq \beta^2 \}.
\]
For a given $A_{\beta}$ the task oriented quality measure $\mu$  is then defined as
\[
\mu(G) = \sup \{\beta \geq 0, \textrm{such that } A_{\beta} \subset G(B^n_1\cap \textrm{FC})\}.
\]
when ${\bf a}=0$ and a slightly different expression holds when ${\bf a}\neq 0$ (see \cite{LiSastryJournalRA} for details). 
Task ellipsoids have pros and cons. The advantage is that their definition is related
to the task and aims at enforcing wrenches that are useful for the task.
 The disadvantage is that, as acknowledged by the authors,
their definition extensively relies on experience and in-depth knowledge of the task
being performed.

With the objective of maximizing $\mu(\cdot)$, the grasp point calculation problem
is then formulated as a constrained optimization problem.  For a given multifingered
grasp $\mathcal{G}$, let $\bar{\mu}$ be the value of the quality measure computed
on the map induced by $\bf G$. Then the problem is
\[
\max_{{\bf c}_1,\dots,{\bf c}_{n_c}} \bar{\mu}
\]
subject to the constraints that the various ${\bf c}_i$ are placed on the boundaries
of the object $\mathcal{B}$ being grasped. The main problem with this formulation is that
this optimization problem needs to be solved numerically and its solution is computationally hard.
These problems are partially mitigated by the method we propose next.\\

Zhu and Wang \cite{ZhuTransactions2003} propose the first algorithm capable of synthesizing a force
closure grasp on a 3D object with an arbitrary number of contact points and
applicable to 3D objects with piecewise smooth surface.
The idea is very simple yet elegant. A numerical test to determine if a set of contacts
achieve force closure is presented. Under mild assumptions the function defining the test
is differentiable with respect to the grasp configuration. Because such derivatives
can be  computed exactly, the space of grasps is then searched using a gradient descent
and a grasp achieving force closure is eventually determined. This method then sidesteps
the heavy numerical optimization problem evidenced in the previous discussion.

The novel force closure test proposed by the authors is based on the so-called $Q$ norm $||\cdot||_Q$,
a special norm different from the commonly used $L_2$ norm. Starting from this special norm (see
\cite{ZhuTransactions2003} for details), two  distances between a point $\bf p$ and a convex
polyhedron  $A$ are defined:

\[
d^+_Q({\bf p},A) = \min_{{\bf a}\in A}||{\bf a}-{\bf p}||_Q
\]

\[
d^-_Q({\bf p},A) = \min_{{\bf a}\in \partial A}||{\bf a}-{\bf p}||_Q.
\]

\noindent For appropriate choices of $Q$ the two distances $d^+_Q$ and $d^-_Q$ can 
be computed in closed form and are differentiable almost everywhere.  
Let $\mathcal{G}$ be a multifingered grasp with $n_c$ contact points ${\bf c}_1,\dots,{\bf c}_{n_c}$.
Note that each ${\bf c}_i$ is uniquely defined once we know $\bf q$, therefore
 we could also write $\mathcal{G}(\bf q)$ or ${\bf c}_i({\bf q})$.
In addition, because of this dependency when a derivative with respect to the contact point
${\bf c}_i$ is considered, one can also consider the derivative with respect to $\bf q$.
Let ${\bf w}_i$ be the wrench associated with every contact point.
Zhu and Wang prove a novel result showing that a necessary condition to achieve force closure is  
$d^+_Q(0,\textrm{co}\{{\bf w}_1,\dots,{\bf w}_n\})$, where $\textrm{co}$ indicates the convex hull.
This condition is necessary but not sufficient. However a necessary and sufficient condition can
be determined using  $d^-_Q$ (details are omitted because this derivation is rather involved).
This test is in essence a test on the sign of $d^-_Q$, i.e., the grasp achieves force closure if
and only if $d^-_Q<0$.
The grasping planning algorithm is then implemented by the following iterative procedure
where $dQ(q)$ is a case-by-case function defined as $dQ^+$ or $dQ^-$.

\begin{algorithm}
Let $\mathcal{G}_0$ be the grasp associated with ${\bf q}_0$\;
\Repeat{$d_Q({\bf q}_k < 0)$}{
${\bf q}_{k+1} = {\bf q}_k - \lambda \frac{d Q({\bf q})}{d{\bf q}}$\;
}
 \caption{Grasp planning based on $Q$ distance}
\end{algorithm}

We conclude this discussion noting that in addition to the objective of obtaining force closure, 
it is also possible to include  constraints to the grasp, like for example
imposing contacts in certain areas of the object being grasped. Moreover it is also possible
to fix the location of some contacts (say the contact positions of two fingers) and let the
algorithm plan the contacts for the remaining $n_c-2$ fingers. Finally, the terminating condition
could be replaced to let the algorithm find a local minimum rather than stopping as soon as $Q$
becomes negative. In that case the algorithm not only determines a force-closure grasp, but
it also optimizes the grasp in terms of resistance offered to  an external load.

The authors conclude the paper showing how the algorithm can be used to plan contact points for various complex objects.\\

When models for the object and the hand are available, one way to plan a grasp is to rely on physical simulation to determine
if a candidate grasp is good or not. 
This approach falls in the {\em geometric only} category because it is purely based on geometric properties and does
not rely on otherwise collected data.
This line of research has been pioneered in Allen's research group  \cite{AllenICRA2004} and relies on the subtle
 distinction that grasp planning is  performed using a simplified object model, but grasp simulation uses a
highly accurate geometric and physical model.
Within this framework, a key intuition is that the great number of possibilities emerging
in grasp planning can be reduced by relying on the concept of {\em pre-shape}, i.e., a fixed number
of general purpose hand configurations. This approach is then coupled with their
GraspIt! simulation environment \cite{Graspit} and aims to be largely hand-independent, as opposed to methods
specifically tailored to a certain class of hands or specific robot hardware. Pre-shapes simplify the complexity on 
the hand side because it restricts the search space. However, the approach performs further simplification also
on the object side, because the geometric model is not assumed to be accurate, but rather an approximation based
on a set of primitive shapes, i.e., spheres,  boxes, cylinders, and cones. Each primitive shape is associated with
a set of grasping strategies. A grasp strategy defines the pre-shape and is parametrized in a way similar to
those presented in Section \ref{sec:grasp_parametrizations}, i.e., in addition to the spread of the finger, the
position and the orientation of the hand are specified, too. At this point each grasp is tested for quality 
using the Ferrari-Canny metric \cite{FerraryCanny}. The evaluation does not take place in the physical world
but is rather performed inside GraspIt! assuming a Barrett Hand mounted on a Puma 560 arm.
The simulation system moves the hand along the approach direction
and eventually closes the fingers, assuming no impediment to the motion emerged. At that point the metric
is computed and the quality of the grasp is returned.

The same principles are further developed in \cite{AllenICRA2007}. The main difference is that rather than relying
on the elementary shapes described above, the authors propose to use a tree of superquadrics (this method
shares many similarities with the system presented in \cite{HubnerICRA2008}  that we discuss  in Section \ref{sec:modelless_vision}).
 The simple idea behind this method is that a set of superquadrics
better approximates the the shape of a complex object than a single superquadric. There is a rich literature in 
computer graphics describing how to ``best" approximate a single shape with a superquadric.  There also exist
methods to compute hierarchical superquadric decompositions where a superquadric is recursively
split until a certain error threshold is met. The authors maintain that this method
performs poorly when used for grasping applications and propose instead to set a priori a number $k$ of
elementary superquadrics, i.e., superquadrics that cannot be further split.  The choice of this parameter
is critical and according to the authors it depends on the dexterity of the hand being used (but not on 
the class of objects being grasped). For the Barrett Hand this number has been determined to be $k=6$.
As for the previous method, a set of tentative canonical grasp is considered for each shape, with the advantage
that in this case all shapes are superquadrics of different size because of the decomposition. After
the set of tentative grasps is generated, the approach resembles \cite{AllenICRA2004}, i.e.,
it uses GraspIt! for simulation and the Ferrari-Canny metric for evaluation purposes.\\

\subsubsection{Data driven algorithms}
Goldfeder  and co-workers proposed a data driven method \cite{ColumbiaGraspDatabase,AllenDataDrivenGrasping} that relies on the  
Columbia Grasp Database. The database, described in the same paper, includes a large number of geometric models
of objects taken from a publically available shape benchmark. Each object is appropriately scaled to fit into a
human or robotic hand, and is further associated with grasp relevant information. In particular, for every object
the database stores three types of information. 1) a {\em pre-grasp}, i.e., the pose of the hand right before the hand contacts the object;
2) the grasp itself, implementing a form closure grasp; 3) quality metrics for the grasp, as defined  in \cite{FerraryCanny}.
Note that grasps were computed for two versions of the Barrett hand (with different friction), and a human hand.
The paper further explains how these grasps were automatically pre-computed using the GraspIt! software and other
preexisting grasping methods, but these details are immaterial to the data-driven grasping method they propose. 
The algorithm
 takes as input $\alpha$, a geometric model of the object to grasp, and returns a set $R$ of tentative grasps for the object. To 
 compute $R$, the algorithm scans the database for the $k$ most similar objects, according to a shape similarity metric.
 Next, for each of the $k$ similar objects, it aligns $\alpha$ with the object and tries to apply the associated 
 precomputed grasps stored in the database. The final  grasp is evaluated using GraspIt! and if the computed quality
 exceeds a given quality threshold $\tau$, it is included in $R$, otherwise it is discarded. \\

The method critically depends on three parameters. The first is the shape similarity metric used to 
 identify similar objects. In the paper the authors use the $L^2$ distance between Zernike descriptors \cite{Zernike}
 combined with a $k$ nearest neighbor approach. 
 The second parameter is $k$, i.e., the number of nearest neighbors to consider. Evidently, in picking $k$ there is
 a tradeoff between speed and accuracy. In the papers the authors fix $k=5$. Finally, the threshold $\tau$ is chosen
 to decide if a computed grasp shall be included in the result set $R$ or not.
 The authors report that the method is capable of computing grasps in about 20 seconds (computational times were given in 2009).
 The performance of the algorithm is compared with a ground truth baseline giving the best match in the database. The analysis 
 provided by the authors show that the selected shape-recall method offers comparable performance. The authors also compare
 the data-driven method with another algorithm they formerly proposed \cite{CiocarlieEigenGrasps} and show that the data-driven approach is 
 faster and more accurate.\\
 
We conclude by observing that this method shares many features with \cite{BalaguerIJHR2011}, although in \cite{BalaguerIJHR2011} we assumed 
the geometric model was acquired on the fly, and then, according to our taxonomy, we implemented a model-less algorithm.

\subsection{Model-less Algorithms}
A significant body of recent research on grasping considers the situation where a geometric
model of the object to be grasped is not available upfront. This problem domain is considered to be
 much more challenging, but it also has a much broader applicability
 than model-based algorithms and presents the opportunity for broader application sets.
 To remedy the lack 
of an \textit{a priori} model, these grasping algorithms rely on sensor data acquired at run time\footnote{We
purposefully ignore sensorless algorithms like \cite{GoldbergAlgorithmica1993,GoldbergAlgorithmica2000} as their scope and 
applicability falls outside the scope of this primer.}.
Typically, these include vision and tactile sensing. Information acquired at run time
could then be used to build  an approximate or incomplete object model on the fly,
or to compute a grasping strategy directly from raw data without attempting to produce a model.
One could retort that an algorithm that builds an approximate model on-the-fly is not
model-less anymore. Here we consider an algorithm model-less if it does not assume the availability
of a model \textit{a priori}. Then, according to this definition, an algorithm building a model on-the-fly
is still model-less. 
It goes without saying that the use of machine learning algorithms is prevalent in this area. 

\subsubsection{Model-less algorithms based on vision}
\label{sec:modelless_vision}
In this section we present two algorithms capable of synthesizing a grasp for an unknown three dimensional 
object based on visual input. While numerous methods were developed for the special case of planar objects,
the three dimensional case is significantly harder and has been considered in its full generality only
recently.\\

\noindent {\bf Methods that construct a model}\\
When a camera system is available, one way to solve the grasping problem is  to construct 
a model of the object and then plan a grasp based on the recovered model. Given the copious amount 
of work in shape reconstruction available in the computer vision literature, it is evident that
many different methods could be engineered around this idea. Therefore, in this primer we will only
analyze a few selected ones. The reader is also referred to \cite{KragicRAS2010} for a more detailed
discussion of this topic.\\

In \cite{DuneIROS2008}, the author proposes a shape estimation method aimed for implementing a {\em ``One click grasping tool"}.
The goal is to extract a geometric model of an object on-the-fly, so that a robotic gripper can then approach and restrain it.
The envisioned application is an assistive device for disabled people, like a robotic arm mounted on a wheel chair.
The underlying assumption is that the object to be grasped is convex (or roughly convex). The shape of the
object is then approximated using a quadric and the parameters of the quadric are identified from multiple views
after the contour of the object has been extracted from the image plane. In order to take images from
different vantage points, it is assumed a camera is mounted on the robotic arm and the authors describe
a method to move the camera around the object (explore) with the objective of minimizing the estimation 
error. Once the estimation process for the parameters of the quadric has been completed, the object
is grasped by aligning the gripper with the minimal dimension while at the same time being orthogonal
to the largest axis.\\

Besides methods trying to estimate shapes starting from images or sequence of images, there is also 
work attempting to achieve the same goal starting from point clouds or stereoimages.
The method presented in \cite{HubnerICRA2008} is similar to the reference \cite{DuneIROS2008} discussed above, 
in the sense that the objective is to find a primitive shape offering a good approximation of the object to
be grasped. The difference is that in \cite{HubnerICRA2008} the input is a point cloud and that rather than 
using a quadric, the authors opt for a hierarchical decomposition based on minimum volume bounding boxes (MVBB).
  The rationale behind this choice is that boxes are 
simple shapes to reason about, but at the same time offer poor approximations of complex shapes resembling
everyday objects that a robot may want to grasp. Therefore a hierarchical decomposition is used in order to
retain the advantages and mitigate the problems. It is worth noting that these hierarchical decompositions
have been extensively used in collision detection algorithms, but with the notable difference that in 
that case a geometric representation of the object based on primitive shapes (e.g., triangular meshes) is
often given, whereas the authors aim at processing point clouds. The authors then propose a
{\em fit and split} algorithm to decide if a bounding box should be split and, if that is the case, where.
The method is tested using GraspIt! \cite{Graspit} and the Columbia Grasp Data Base \cite{ColumbiaGraspDatabase}
using a five finger robotic hand. The objective is to evaluate how the proposed method performs in terms
of identifying force-closure grasps. The grasping strategy follows the method given by Pelossof et al. \cite{AllenICRA2004}
and formerly described.\\

\noindent {\bf Methods that operate in feature space}\\
The first algorithm capable of grasping a previously unseen object was presented by Saxena et al. \cite{SaxenaNgIJRR}. 
The algorithm was implemented and tested on a Barrett arm equipped with the Barrett hand for tasks like unloading 
a dish-washer. 

The algorithm processes a single image of an object to be grasped and identifies
in the image the pixels corresponding to good grasping points. This information is then
passed to a robot controller that will attempt to grasp the object by placing the fingers
in the locations corresponding to the identified grasping points.
The training stage of the algorithm relies on a large set of images of objects to be
grasped where good grasping locations have already been identified and marked (e.g., the
handle of a mug). Interestingly, the authors use a database of synthetic images that is
publicly available. During the training stage,
 a high-dimension feature vector
for every pixel in the image is created, 
and  then  a classifier discriminating good grasping points from bad grasping points
is trained. For every pixel, a feature vector is created using various image filters computed 
in a $5\times 5$ neighborhood of the pixel. Each feature $x_i$ is a vector\footnote{Seventeen 
filters are applied on each neighboring pixel in a 5$\times$5 path and on the pixel 
on the scaled version of the image in three different sizes, hence 17*24 + 17*3 = 459.} in $\mathbb{R}^{459}$. 
Since every pixel in the training images is already labeled as good/bad, each feature 
vector $x_i$ can then also be associated with the corresponding binary label $z_i$.
The  classifier is based on logistic regression and works as follows. First, 
a parameter vector $\theta^* \in \mathbb{R}^{459}$ is learned through maximum likelihood:
\[
\theta^* = \arg \max_{\theta} \prod_i \Pr(z_i|x_i;\theta).
\]
Classification at run time then works as follows. First, the feature vector $x$ is computed
for every pixel in the image of the object to be grasped. Then, the probability that
such pixel corresponds to a good grasping point (i.e., $z_i=1$) is computed via logistic regression:
\begin{equation}
\Pr(z_i = 1|x;\theta^*) = \frac{1}{1+e^{-x^T\theta^*}}.
\label{eq:log_reg}
\end{equation}
This step assigns a probability to every pixel and pixels can then be sorted accordingly
to these probabilities. The identified pixels can then be used to guide the robot hand
and complete the grasp maneuver. In the original paper, this  is done by using a 
stereo camera providing depth information, but various options are possible.

The notable part of this algorithm is that at run time, the 
robot is not bound to grasp objects it saw during the training stage, but can rather
handle arbitrary objects. Despite its merits, this algorithm has two drawbacks.
The minor one is that computing $\theta^*$ is very time consuming. Although
this step is off-line, it takes time to retrain the algorithm when new images 
have to be included in the training set. The major problem is that at run time, 
a high-dimension feature vector has to be computed for every pixel in the image 
in order to then compute Eq. \ref{eq:log_reg} for every pixel. For large images ,
this process become very time consuming and the frame rate is significantly impacted.
It should be noted that, as evidenced in the accompanying videos available on the authors' 
websites, the robot largely relies on pinch grasps or grasps where the whole hand
envelops the object. This approach works well in practice, but does not strictly 
follow the force/form closure taxonomy usually used in robotic grasping.

Balaguer and Carpin later on showed that the algorithm maintains a good performance
even when using a much smaller feature vector \cite{BenICRA2010}, i.e. 54 features rather
than 459. This reduction is possible by using a smaller window around the pixel, and just
a subset of the original image filters. The advantage of this method is that every image
can then be processed in a much shorter time (a speedup of a factor of 8 was reported).\\

Koostra and colleagues \cite{KragicIJRR2012} also propose a system to grasp unknown objects 
based on vision.
The system, called Early Cognitive
Vision (ECV), extracts three-dimensional features that are then used to compute grasps with
two and three fingers. The system works as follows. Two types of features are extracted, namely from
edges and surfaces in the image.
Edges are extracted from the left and right images and grouped into groups like contours.
Groups from the left and right image are then matched to each other.  
Surfaces in the images 
 generate three different types of features that the authors call {\em texlets}, {\em surflings},
and {\em surfaces}. The idea behind these features is to generate a hierarchy of features incorporating
information at higher and higher levels of abstraction.
They represent local properties of a surface with texture information (texlets), which are at the bottom of the hierarachy. 
Each texlet includes the center of the patch ${\bf p}$, the normal ${\bf n}$, and the color $c$.
To build a hierarchical representation, a clustering algorithm is then applied to
texlets to obtain the next level of surfling features. As a collection of local planar patches, a surfling
represents a rectangular planar area.  Finally, surflings are also grouped together in the highest 
level of the hierarchy, i.e., surfaces. As the name suggests, {\em surfaces} aim to represent surfaces
of objects, and color is therefore not included to account for regions with different colors.\\

Once edge and surface features are extracted, a grasp is synthesized for the special cases of
two and three fingers.  Rather than relying on a database of predetermined associations between objects
and grasps, the authors consider a set of basic grasps that are associated with specific geometric attributes.
The correct grasp for an object is then determined by analyzing the previously described features. 
Three different grasps are considered, namely an encompassing grasp,
a top pinch grasp and side pinch grasp. The positioning of the fingers to implement this grasp is
then determined by processing the extracted contour and surface features.\\

The method has been demonstrated both in simulation and on a real world system using a Schunk parallel
gripper and a Schunk Dexterous Hand (three fingers). A BumbleBee 2 stereo camera was used for image acquisition.
The experimental setup considers various different objects. By definition, and in contrast to the previous 
method that required a training stage, all these objects are novel. Emphasis is given on 
grasping performance in the presence of clutter, but no information is provided regarding the 
computational complexity or the time needed to compute these grasps.

\subsubsection{Model-less algorithms based on tactile sensing}

When a model for the object is not available, a grasp can be synthesized  using touch sensors.
In this case, a robotic hand equipped with tactile sensors can acquire information about the object being 
grasped before actually attempting a grasp. Similar to vision-based approaches, some tactile-based algorithms reconstruct a model of the object, whereas others do not. Note that these methods to not combine tactile data
with vision data, but rather rely only on touch.
Given the still relatively scarce availability of 
 affordable, reliable tactile sensors, this line of research is rather novel and there are 
 much fewer contributions than systems relying on visual input.\\

\noindent {\bf Methods that construct a model}\\
Bierbaum and collaborators have used touch sensors to infer a model of the object to be grasped \cite{BierbaumTactile2008,BierbaumTactile}.
In particular, in \cite{BierbaumTactile2008} they maintain to be the first to have developed a method for tactile
exploration using a multifingered robotic hand, in contrast to previous methods strictly relying on a single finger for exploration.\\

The objective of the presented system is to reconstruct the 3D shape of an unknown object by touching its surface using the fingers.
It is assumed that just the fingertips are equipped with tactile sensors, and  that prior information
about size, position, and orientation of the object is also provided.
The key problem to be addressed is how to move the fingers in order to explore the unknown shape of
the object, a problem related to  mobile robot exploration. Exploiting this similarity, the authors embrace a method
based on potential fields associated with a set of points generating attractive and repulsive forces. 
Initially, using the prior information about the object, a uniform three-dimensional grid is placed around the object.
An attraction point is then created inside every grid cell, and exploration starts based on the associated attractive field.
Whenever a finger crosses a cell without making any contact, the attractive point located inside the cell is eliminated.
When a finger detects a collision inside a cell, a repulsive point is generated. The exploration of the 
surface of the object occurs then by following the negated gradient of the potential field function. As with every method based
on potential fields, the main issue to consider is the possibility of getting stuck in a local minima. To cope with this
problem a reconfiguration policy is added. Reconfiguration occurs when a deadlock situation is detected, i.e., when the hand
has not been moving for a while. When this condition is triggered, the role of attractive and repulsive points is temporarily 
swapped, so that the hand moves away from the object and is biased towards an unexplored part of the space. While in \cite{BierbaumTactile2008} 
the authors just describe the exploration stage, i.e., how a 3D model for the object can be built, in \cite{BierbaumTactile}
a grasping strategy is presented, too. The main challenge to face is that the 3D shape reconstructed during the exploration 
stage is inherently discrete, because of the aforementioned grid. This obstacle is overcome using the
method presented in \cite{PertinTroccaz} where geometric features are extracted and grasping is executed trying to place the fingers
on faces offering a sufficiently high grasping score.
The method is tested on a simulation system replicating the 6-DoF Mitsubishi RM 501 robotic arm equipped with a FRH 4 Hand, a hand with
9-DoFs and four fingers.\\

An alternative approach was proposed by Platt et al. \cite{GrupenTRO2010} whereby the system does not need a
perfectly accurate description of the object geometry. Instead, through an iterative refinement process, an initial
grasp is updated until force closure is achieved. The refinement process is guided by a force feedback system.
The system could then be used jointly with a traditional planner operating on an incomplete or noisy model. The set
of contact points provided by the planner could be inaccurate and/or not guarantee force closure, but would
provide a good starting point for the iterative process implemented by the controller. Specifically, the controller
implements a null-space approach where residuals for force and moments are combined to achieve force closure.
This method was tested on a whole arm manipulator (WAM) arm equipped with a Barrett hand retrofitted with load cells located at the fingertips
for force feedback.\\

\noindent {\bf Methods that operate in feature space}\\
 Dang et al.  use the term  {\em blind grasping} to indicate methods where the
 robot relies on tactile sensors only, but does not use vision  \cite{AllenICRA2011,AllenLearningGraspStability}. 
 Their method is feature based and does not attempt to use tactile data to reconstruct an approximate model of the object.
The method assumes the availability of a grid of tactile sensors placed on the fingers.  
It is assumed the sensors are subdivided into a certain number of pads (e.g., 4), but 
this assumption is not critical for the method.
When performing a grasp, each tactile element in the grid will be either inactive (i.e., no contact)
or active. Note that active tactile elements return the actual force, and not just a binary 
indication of whether they are activated.
Since the position of the tactile sensors is fixed on the hand, assuming that 
the joint values are known, it is then possible to determine
the location of the contact points through forward kinematics. Starting from these premises, the grasping method
has some ideas in common with the work of Saxena \cite{SaxenaNgIJRR} we described in
 Section \ref{sec:modelless_vision}. In fact, the method also relies on a training based on a set of labeled grasps
 (good/bad), it
 computes a feature vector at run time, and uses a classifier to decide if a grasp is good or bad
 based on the information gathered during training.
The details are described in the following. During the training stage a large number of grasps is
considered, and for each of them the set of contact points is recorded. To reduce the dimensionality
of this dataset, just the position of the contact points  is recorded (information about the direction
of the force is discarded), and points are then clustered using $K$ means. This reduced 
dataset is called {\em contact dictionary} in order to evidence a similarity with methods based
on the Bag of Words principle \cite{BoW}. The dictionary is then
\[
D = [{\bf \hat c}_1,{\bf \hat c}_2,\dots,{\bf \hat c}_p]
\]
where each of the ${\bf \hat c}_i$ is a contact point in $\mathbb{R}^3$. Note that these are not 
actual contact points, but rather the results of the clustering process. The authors report
that $p=64$ was used in their experiments and provided a good tradeoff between speed and accuracy.

When the robot hand makes contact with an object, a vector of $q$ contact points 
is generated, i.e., a grasp $\mathcal{G}$ is represented by a vector 
\[
\mathcal{G} = [{\bf c}_1,{\bf c}_2,\dots,{\bf c}_q].
\]
In general, $p \neq q$.
Starting from $D$ and $\mathcal{G}$ a $p$-dimensional feature vector is computed relating the grasp $\mathcal{G}$ to
the dictionary $D$. The feature vector is defined as
\[
F(\mathcal{G},D) =\sum_{i=1}^q H({\bf c}_i,D) \frac{f_{{\bf c}_i}}{S_{{\bf c}_i}}
\]
where $f_{{\bf c}_i}$ is the force sensed at contact point ${\bf c}_i$ and $S_{{\bf c}_i}$ is
the total force sensed. The term $H({\bf c}_i,D)$ is a $p$-dimensional vector measuring how ${\bf c}_i$
relates to the dictionary. To compute its $j$-th component one can use the following function
\[
h_j = \exp \left( - \frac{|| {\bf c}_i-{\bf \hat c}_j||^2}{\sigma^2}\right).
\]

The training stage is completed using the contact information of more than 24000 grasps from the Columbia Grasp
Data Base \cite{ColumbiaGraspDatabase}. For each grasp the feature vector is computed and 
and associated with the corresponding binary label (good/bad grasp). Then, a support vector machine
(SVM) is trained based on this data to discriminate between the two classes. Emphasis is given
to reduce the false positive rate, as the authors speculate that from a
practical point of view this kind of error is more  severe than a false negative (i.e., not executing
a good grasp because the classifier erroneously considers it to be a bad grasp.) At run time the robot approaches an object
lying on a plane
and grasps it with a tentative grasp $\hat{\mathcal{G}}$. The corresponding feature vector $F(\hat{ \mathcal{G}},D)$
is computed and then classified using the SVM. If the SVM classifies  $\hat{\mathcal{G}}$ as a good grasp,
the robot tries lifting the object, otherwise it tries a different grasp.\\

The method is validated using a Barrett Hand with a good performance. However, there are some drawbacks,
too. For example, in order to initially make the tentative grasp $\hat{\mathcal{G}}$ the robot needs 
prior information about the location of the object. If the position is not available
or if is noisy, the method will be either inapplicable or doomed to fail. 

\section{Dual Arm Grasping}

Various technological developments and practical reasons continue to drive research in 
dual arm grasping and manipulation. From a practical point of view, certain
grasping tasks require  interaction with larger sized parts that may have physical
properties (weights, inertias .. ) and manipulation requirements that would require two or more robotic arms utilizing  simple grippers or even possibly robotic hands. In these scenarios, the use of two arms is then preferred.
Moreover, there is an increasing expectation that robots will work side-by-side
with humans in a variety of everyday tasks ({\em corobots}). In this case robots
have to be able to use the same tools and operate in the same environments designed
for humans. As these designs often assume end-users are equipped with two hands, 
a dual arm design will then ease the integration of robots into these environments.
Finally,
continued research in humanoid robotics is naturally linked to dual arm 
grasping and manipulation.
Recent surveys on the topic appeared in \cite{KragicDualArm} and \cite{DualArmHandbook}.\\

From a high-level point of view, dual arm grasping can be treated with the same formalism 
used for single arm robots. In fact, one can think of the dual arm robot as just one system
with a higher number of degrees of freedom. According to a taxonomy proposed in \cite{KragicDualArm},
this paradigm is called {\em coordinated} operation, and it refers to the situation where
the two arms perform different parts of the same task\footnote{This is opposed to what
they call {\em non-coordinated} operation, where the two arms perform separate tasks and can
therefore be treated separately.}. From a grasping perspective, then, one can define extended
versions for the grasping matrix  $\bf G$ and the Jacobian matrix $\bf J$ and then perform
the same analyses presented in Section \ref{sec:formalism} for assessing form and force
closure.

\section{Grasp Performance Benchmarking}
Standardized performance testing, an emerging tool within the robotics community, is proving itself worthy within other robotics community disciplines and offers the benefits of an ``honest broker" to measure how well a system performs in a particular ability \cite{Bonsignorio2014}.  Ultimately, physical measurements assess grasping performance using measurement techniques external to the system under evaluation. The results of such evaluations and benchmarks help to match capabilities to end-user needs, as well as to help developers improve their product designs. Section 5.1 discusses existing grasp measures and more abstract grasp properties for robotic hands, tactile sensing, and grasp algorithms within the robotic grasping community. Section 5.2 takes an application approach to overview the grasping problem in order to fully understand the scope of measurements. Finally in section 5.4, the results of our literature review and task analysis are used to formulate some concepts towards a set of performance evaluation and benchmarking techniques for automated grasping systems.

\subsection{Quantitative Grasp Measures}
The physical results of grasping are reported using both qualitative and quantitative data. Qualitative data is a categorical measurement expressed by means of a natural language description where quantitative data is a numerical measurement. Qualitative measures expressing the ability to grasp an object are commonplace and typically use pass/fail indicators along with a description of how well a grasp was performed on a given test (e.g., grasp A is not as stable as grasp B,  the object was ejected from the grasp).  Another aspect of performance testing is functional vs. non-functional testing. Functional tests evaluate a robotic hand and overall robotic system's ability to perform the grasp required to accomplish a specific task (e.g., holding and operating tools, grasping and turning valves, and operating a door knob after unlocking it with a key) \cite{DARPARC1,DARPAAC,NISTASTM},  while non-functional tests would be designed to measure more general properties of a robot hand outside the scope of an integrated robotic system. Both qualitative and quantitative measures can be used to express the results of functional and non-functional tests. Qualitative measures are easily found in robotic grasping research literature; however, examples of applying quantitative measures to evaluate grasp performance are sparse.

\subsubsection{Volumetric}
\begin{figure}
\centering
\includegraphics[width=12cm]{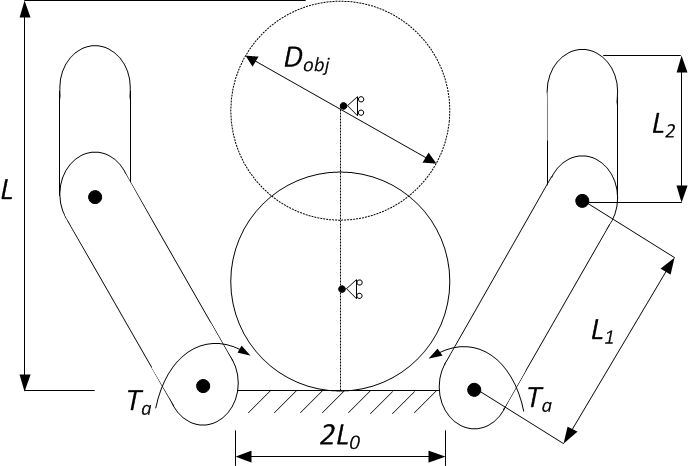}
\caption{Reproduced from \cite{Kragten2010}, initial positions of a freely moving cylindrical object with respect to the palm of a hand to determine the ability to successfully grasp this object. Palm position is represented by the solid object and the pinch position by the dashed object. $D_{obj}$ is the diameter of the object, $2L_0$ is the width of the palm, $L_1$ is the length of the proximal phalanx, $L_2$ is the length of the distal phalanx, $L$ is the length of the finger. A torque $T_a$ applies to the base of the fingers.}
\label{fig:volume_test}
\end{figure}

The authors of \cite{Kragten2010} propose a benchmark to measure the kinematic ability of a robotic hand to grasp objects. In particular, cylindrical objects were used of increasing diameter (40, 50, 55, 60, 63, 75, 90, 110, 115, and 120 mm) under both pinch grasps and enveloping grasps. In the case of a pinch grasp, the outermost point of the object circumference was placed a distance of L from the palm, and in the case of an enveloping grasp, the object was placed against the palm. Also noted during grasps were the cases where a pinch grasp resulted in the hand pulling the artifact towards the palm resulting in a transition to a final enveloping grasp equilibrium. A performance metric was defined as follows:

\[
 Q_{grasp} = 
\frac{\frac{\pi}{2}\Delta D_{obj}}{2L+2L_O}
\]

\noindent where $\Delta D_{obj}$ is the difference between the diameter of the largest and smallest graspable object, $L=L_1+ L_2$ is the length of one finger, and $2L_0$ is the palm width of the hand (Figure \ref{fig:volume_test}).

\subsubsection{Internal Force}
\begin{figure}
\centering
\includegraphics[width=12cm]{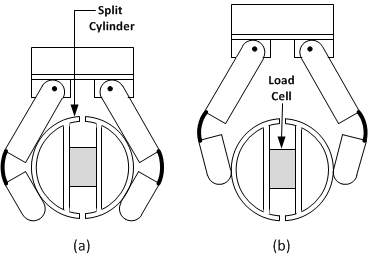}
\caption{Reproduced from \cite{Odhner2014} split ring test apparatus to measure power grasp (a) and finger-tip force (b).}
\label{fig:force_test}
\end{figure}

Odhner et al. implemented a test apparatus to test the power grasp capabilities of the iRobot-Harvard-Yale (iHY) Hand, a compliant under-articulated hand that used tendons to actuate finger motion \cite{Odhner2014}.  The apparatus was constructed of a split cylinder and a load cell attached at the cylinder center. The apparatus was oriented such that it was symmetric with the fingers and the load cell measured the force exerted between the opposing fingers in the direction of the split. The same test artifact was used to measure both power grasping and finger-tip grasping (Figure \ref{fig:force_test}). \newline

Romano et al. presents quantitative testing when evaluating the performance of a novel robotic grasp controller \cite{Romano2011}. The system's ability to control delicate manipulation tasks was evaluated with crushing measures. Crushing was defined as a deformation of 10 mm beyond the initial surface contact. There was no indication as to how these measures were made.

\subsubsection{Resistance to Force and Slip}
\begin{figure}
\centering
\includegraphics[width=12cm]{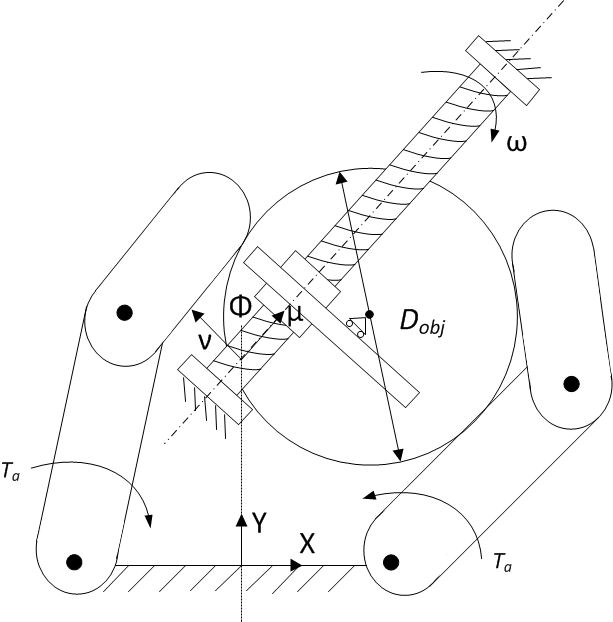}
\caption{Reproduced from \cite{Kragten2010}, a schematic for the setup where a cylindrical object with a diameter $D_{obj}$ is pulled out of the hand at a constant slow speed $\omega$ in the direction of $\mu$ while the fingers are at a constant torque $T_a$. The object is free to move in the direction $\nu$, which is perpendicular to $\mu$. The resultant of the contact forces on the object in the pull direction $\mu$ is measured.}
\label{fig:slip_test}
\end{figure}

A benchmark in \cite{Kragten2010} tests the ability to hold objects. Again using cylindrical objects, the object is placed against the palm of the hand and grasped. The object is slowly moved along a straight line (5 mm/s) in a disturbance direction $\mu$, with the object allowed to move in the perpendicular direction $\nu$ (Figure \ref{fig:slip_test}). The force is measured throughout the pulling direction over several pull directions $\Phi$ and the maximum pull force is recorded for each. A performance metric was designed as follows:

\[
 Q_{hold} = 
\frac{FL}{T_a}
\]

\noindent where $F$ is the minimum force needed to pull an object out of the hand, $T_a$ is the constant actuation torque applied at the base of the fingers, and L is the total length of the finger. In \cite{Meijneke2011} the authors conduct similar experiments with grasps as in \cite{Kragten2010}, but also independently measure the contact force using a load cell internal to the cylinder and coupled to the hand through a ball bearing protruding through a hole in the cylinder. In these tests, the authors are relating the forces exerted on the cylinder by the hand to the forces required to pull the cylinders from the hand. \newline

Romano et al. test a robotic hand's ability to control delicate manipulation tasks using slippage measures \cite{Romano2011}. Slippage was defined was defined in two forms: translation (greater than 5 mm), and rotation (greater than 10 degrees). There was no indication as to how these measures were made. Slip measures were used to evaluate the grasp controller's ability to adjust the minimum grip force necessary to lift an object 10 mm from a table surface and to evaluate the controller's slip response. To evaluate slip response, a cup was stably grasped at a fixed load of 5 N. The cup was loaded incrementally with batches of 15 marbles (about 0.6 N per batch) and the gripper was shaken for two seconds while the cup was observed for slip. The authors also used pressure-sensitive film to capture the forces imposed on an object during placement onto a flat surface.

\subsubsection{Touch Sensitivity}
Dollar et al. presents an experimental setup to test grasp improvements achieved when integrating piezofilm contact sensors with a reactive control algorithm onto the Shape Deposition Manufactured (SDM) Hand \cite{Dollar2010}. The experimental setup consists of a shape artifact constrained to a six-axis force-torque sensor. A nominal grasp pose relative to the position of the artifact to be grasped is determined. Error offsets are then applied to the nominal pose and the forces associated with and without sensor feedback are measured. In addition to the force measurements, a qualitative assessment is applied to measure the success of the grasp.  A successful grasp was defined as one where the object was able to be successfully lifted out of the force sensor mount without slipping out of the hand. Grasp success and contact force data were evaluated at 10 mm error increments from the nominal position. Results indicated that the addition of feedback from the contact sensors on the hand decreases the forces applied to the object during the grasp and increases the range of acceptable positioning offsets that still result in a successful grasp. \newline

Based on Dollar's work, SynTouch LLC reports an experiment for comparing the sensitivity of grasping using tactile sensing technologies\footnote{ This work was performed as part of a NIST Small Business Innovative Research (SBIR) contract no. DB1341-13-SE-0300 titled Advanced Tactile Sensing Technology for Robotic Hands.}. Using a spherical object fixed to a force plate, the experiment measures the unbalanced forces acting on the object upon making grasp point contacts. The tests were conducted over a range of closing velocities and varied the position of the object to test how grasps can adapt to positional errors. Results showed higher forces with increasing closing velocities and decreasing sensor compliance and were attributed to the speed of the hand's force control loop using the integrated sensor system. The research also presented a mechanism for using the collected data to determine the range of velocities and position errors a robotic hand system can tolerate for a given peak force.

\subsubsection{Compliance}
\begin{figure}
\centering
\includegraphics[width=12cm]{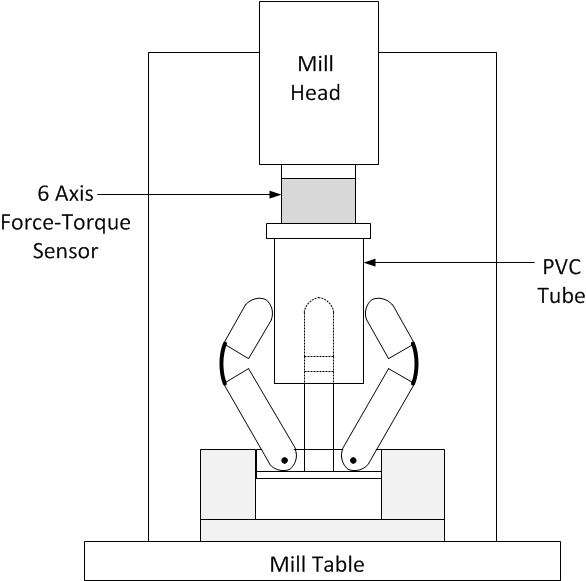}
\caption{Reproduced from \cite{Odhner2014}, experimental setup for measuring stiffness properties of compliant hand.}
\label{fig:stiffness_test}
\end{figure}

The developers of the iHY also developed a test for measuring the compliance of planar and spherical pinch grasps \cite{Odhner2014}. This was accomplished by mounting a 6-axis force-torque sensor to a mill headstock with the iHY hand fixtured in the mill's vice (Figure \ref{fig:stiffness_test}). Disturbance displacements were applied using the three linear axes of the milling machine, and the resultant forces were recorded relative to displacement. Stiffness values were determined by averaging out hysteresis due to tendon friction and the viscoelasticity of polymer pads and flexures by averaging values in both directions of each motion over several cycles.

A linear least squares estimation was used to fit the parameters of a symmetric stiffness matrix K to the data for both the opposed and spherical fingertip grasps.

\[
K=
\left( \begin{array}{ccc}
K_{xx} & K_{xy} & K_{xz} \\
K_{yx} & K_{yy} & K_{yz} \\
K_{zx} & K_{zy} & K_{zz}
\end{array} \right)
\] 

\subsubsection{In-Hand Manipulation}
Odhner et al. reports on experiments used to evaluate the in-hand manipulation capabilities using two fingers of an under-actuated robotic hand \cite{Odhoner2013}. Using several small objects having different width and radius of curvature, manipulation tests were conducted by tracking the position of the object relative to an initial fingertip grasped position. Objects were tracked in six degree-of-freedom space using a commercially available tracking system and the degree of slip was detected by measuring the error between the nominal start and finish object positions after returning the robotic hand fingers to their original fingertip grasp position.

\subsubsection{Grasp Properties and Quantitative Measures}
Grasp synthesis identifies the physical and mechanical properties of grasps and the creation of suitable parameters to quantify them.  In a review of grasp synthesis algorithms, Shimoga identified the main properties of grasping as disturbance resistance, dexterity, equilibrium, stability, and dynamic behavior \cite{ShimogaSurvey}. A grasp with good disturbance resistance can withstand disturbances in any direction. This can be accomplished by form-closure (complete kinematical restraint) where a set of grasp points results in finger positioning that constrains an object, or force closure where grasp point forces applied by fingers constrain motion of the object (specific measures for these are proposed by many researchers and are discussed in Section 5.3). A grasp is considered dexterous if the kinematic properties of the robotic hand allow the object to be moved using a controlled and stable method, a concept also referred to as in-hand manipulation. A grasp is in equilibrium if resultant forces and torques applied to an object by finger and external forces equate to zero. A grasp is considered to be stable if any positional errors caused by external forces in finger or object position can be eliminated once the disturbance is removed. Finally, the dynamic behavior of a grasp is defined as the time response of the grasp for changes in its motion or force trajectories.

Similar to Shimoga, Cutkosky presents the properties of force closure, form closure, stability, and manipulability as analytical measures used to describe a grasp \cite{Cutkosky1989}. Cutkosky also presents internal forces, slip resistance, compliance, and connectivity. Internal forces apply to the magnitude and variance of forces that a hand applies to an object while maintaining grasp equilibrium that is described above. Slip resistance is the magnitude of the forces and moments on the object at the onset of slip. The resistance to slipping depends on the configuration of the grasp, on the types of contacts, and on the friction between the object and the fingertips. Compliance (inverse of stiffness) of the grasped object with respect to the hand is a function of grasp configuration, joint actuation, and structural compliances in the links, joints, and fingertips. Finally, connectivity is the number of degrees of freedom between the grasped object with respect to the hand.

Another important property is indicated in \cite{Dollar2010} that we will call grasp sensitivity, the ability of a grasp to conform to deviations in nominal object position without disturbing actual object location prior to achieving form closure with the object. Grasp sensitivity is a property of a force or contact sensing and associated control algorithms that occurs when achieving form closure. Table \ref{table:grasp_properties} consolidates these measures and maps them to the quantitative experimental methods as described in section 5.1. As indicated, there are multiple performance tests that can be used to assess a given measure; and some measures can be supported using several of the experimental methods found in the literature.

\begin{table}
\centering
\caption{Grasp properties and applicable performance tests}
\vspace{.5cm}
    \begin{tabular}{ | l | p{7.5cm} | p{3.5cm} |}
    \hline \textbf{Grasp Property} & \textbf{Description} & \textbf{Applicable Tests} (from Section 5.1) \\
    \hline Form Closure & Ability to spatially constrain an object from moving when the finger joints are locked when assuming contact between the fingers and the object. & 5.1.1 \\
    \hline Force Closure & There exists a conical combination of contact forces at the points of contact such that any external wrench applied to the object can be resisted. & 5.1.2, 5.1.3 \\
    \hline Dexterity/Manipulability & The ability of the fingers to impart motions to the object using the kinematic properties of the robotic hand allowing the object to be moved using a controlled and stable method. Also called in-hand manipulation. & 5.1.6 \\
    \hline Equilibrium & Resultant forces and torques applied to an object by finger and external forces equate to zero. & 5.1.2, 5.1.3 \\
    \hline Stability & Ability of the grasp to return to its initial configuration after being disturbed by an external force or moment. & 5.1.2, 5.1.3 \\
    \hline Dynamic Behavior & The time response of the grasp for changes in its motion or force trajectories. & 5.1.2, 5.1.3, 5.1.4 \\ 
    \hline Internal Forces & Magnitude and variance of internal grasp forces that a hand applies to an object without disturbing the grasp equilibrium. & 5.1.2 \\
    \hline Slip Resistance & Magnitude of the forces and moments on the object at the onset of slip. & 5.1.2, 5.1.3 \\
    \hline Compliance & The effective compliance of the grasped object with respect to the hand. & 5.1.5 \\
    \hline Connectivity & Number of degrees of freedom between the grasped object and the hand. & 5.1.6 \\
    \hline Sensitivity & Ability to conform to deviations in nominal object position without disturbing actual object location prior to achieving final grasp. & 5.1.4 \\
    \hline
    \end{tabular}
    \label{table:grasp_properties}
\end{table}

\subsection{Analysis of a Grasping Task}
\begin{figure}
\centering
\includegraphics[width=12cm]{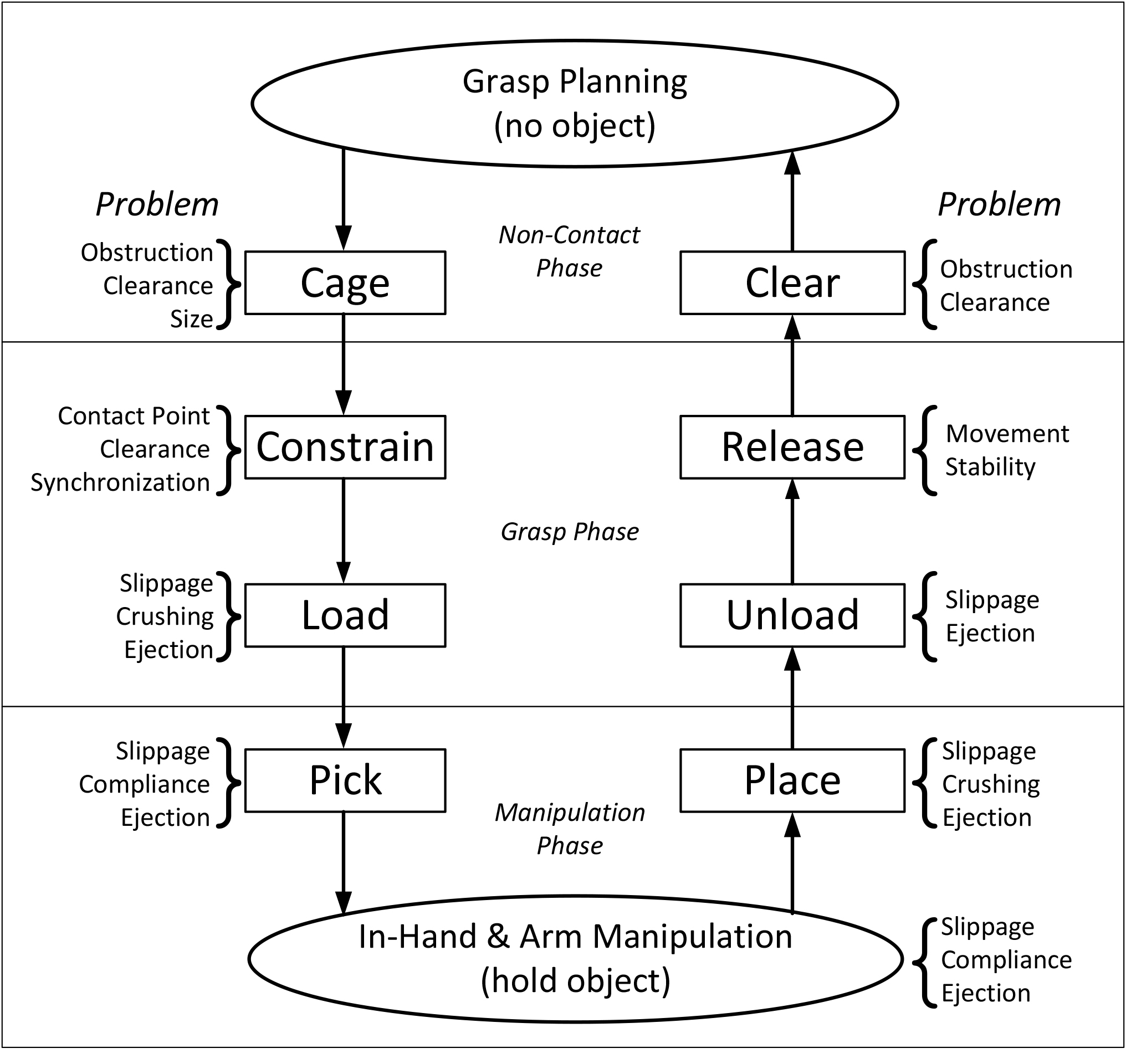}
\caption{Pick and place task segmentation and transitions between grasped and un-grasped states (from \cite{RAS2015}.)}
\label{fig:grasp_task}
\end{figure}

As always, breaking down a problem into its parts can provide novel insights towards its solution. In particular, consider the underlying tasks associated with a robotic pick and place operation for a fully integrated multi-fingered robotic hand (see Figure \ref{fig:grasp_task}). Each task in this particular operation\footnote{ This discussion presents the results of an informal grasp metrics working group organized by NIST as well as grasp metrics and test methods research being performed at NIST. More information at http://www.nist.gov/el/isd/grasp.cfm and http://rhgm.org/. See also \cite{RAS2015}.} possesses a number of associated problems that can serve as a basis for extracting performance measures. More specifically, quantifying the performance of a system in handling these problems can help guide and justify the various strategies taken. Furthermore, identifying the significance of particular performance measures towards different grasping tasks would provide valuable knowledge on necessary functionalities and their performance towards task completion. For example, picking up a part and tossing it into a bin requires minimal position accuracy of the grasped object once the grasp component is completed where picking up a part and performing an assembly operation requires much more accurate positioning throughout the task. Thus, quantifying and suggesting a minimal level of performance in the system's ability to control and measure object position in the latter scenario would be critical in predicting operation success.

A plausible outline for the pick and place operation begins with a "best" set of grasp points as determined from a grasp planner. The hand is positioned by a robotic arm to cage an object by establishing an approach trajectory and offsets that are based on the grasp planning stage. During the cage task segment, it is possible for components of the robotic hand to run into obstructions in the vicinity of the object or run into the candidate object to be grasped due to inadequate \textit{clearances}.

During the \textit{constrain} task segment, the object is spatially confined by the grasp at the grasp points. Sensor-less contacts depend on the positioning accuracy of the hand delivery system and/or the \textit{synchronization} of fingertip position in time. Unsynchronized contact requires minimal force contacts to minimize disturbance to the part, if maintaining part position is important, and requires hand sensing capabilities such as tactile or current sensing. Position problems can occur during this task segment that result in a missed \textit{contact point} where the part is not fully constrained and contact movement due to synchronization issues or inadvertent contact by the hand due to \textit{clearance} issues.  Clearance issues may also result if the object is too small to be grasped as in the case of a three fingered radial grasp on a cylinder where closing the fingers results in a collision between fingers before making contact with the object.

The load task segment applies the calculated forces required to keep a firm grasp on the object. These forces are most often calculated to obtain an efficient grasp based on the forces required to stabilize the object in the presence of gravitational and inertial forces.  Problems during this segment are due to the uncertainty in the kinetics of the system (i.e., object mass), disturbance forces, and torques applied to the object. Uncertainty in the system can lead to the occurrence of \textit{slippage}, damage to the part (\textit{crushing}), or \textit{ejection} when trying to achieve efficient grasp forces.

The pick task segment lifts the object for manipulation. Problems during this segment are due to the errors between gravitational forces and the forces applied during the load task and again result in \textit{slippage}, \textit{ejection}, or \textit{crushing}. Other considerations are variations in object position relative to the robot hand coordinate frame upon picking up the part due to the \textit{compliance} properties of the robot hand.

During the manipulation phase, the part is picked (or lifted) from its grasped position, moved along a trajectory, and placed in a final position. The trajectory could be induced by the robot carrying the hand as well as the robotic hand itself, often referred to as in-hand manipulation. Problems are due to change in part position relative to the hand caused by external forces associated with contacts between the object and the environment throughout the manipulation process, as well as uncertainty in the objects kinetic properties. In addition, fluctuations in the mass of the object as well as exogenous disturbances can occur due to an intermediate assembly operation on the object. All of these force changes can result in \textit{slippage}, \textit{ejection}, or \textit{crushing}.

During the place task segment the object errors are dependent on object positional placement accuracy. In the most lenient case, the object is placed into a bin at a random orientation. In another case the object is positioned on a flat surface where accuracy errors could result in unexpected surface/object contact forces leading to \textit{ejection}, \textit{slip}, or \textit{crushing}. The most complex case is that of assembly, where assembly algorithms are dependent on the positional accuracy of initial contact between the object and the subassembly and the object is subjected to a multitude of external forces throughout the assembly process also leading to \textit{ejection}, \textit{slip}, or \textit{crushing}.

Post manipulation phase, the remaining task segments are a reversal of the task segments that led up to the manipulation phase and the errors associated with these are similar to their counterparts. Here the object is unloaded to the point of zero force contact and released so that the robot hand components clear the object, allowing the hand to be moved to the next operation.

\subsection{Towards Standardized Benchmarks for Robotic Hands}
\begin{figure}[htb]
\centering
\includegraphics[width=10cm]{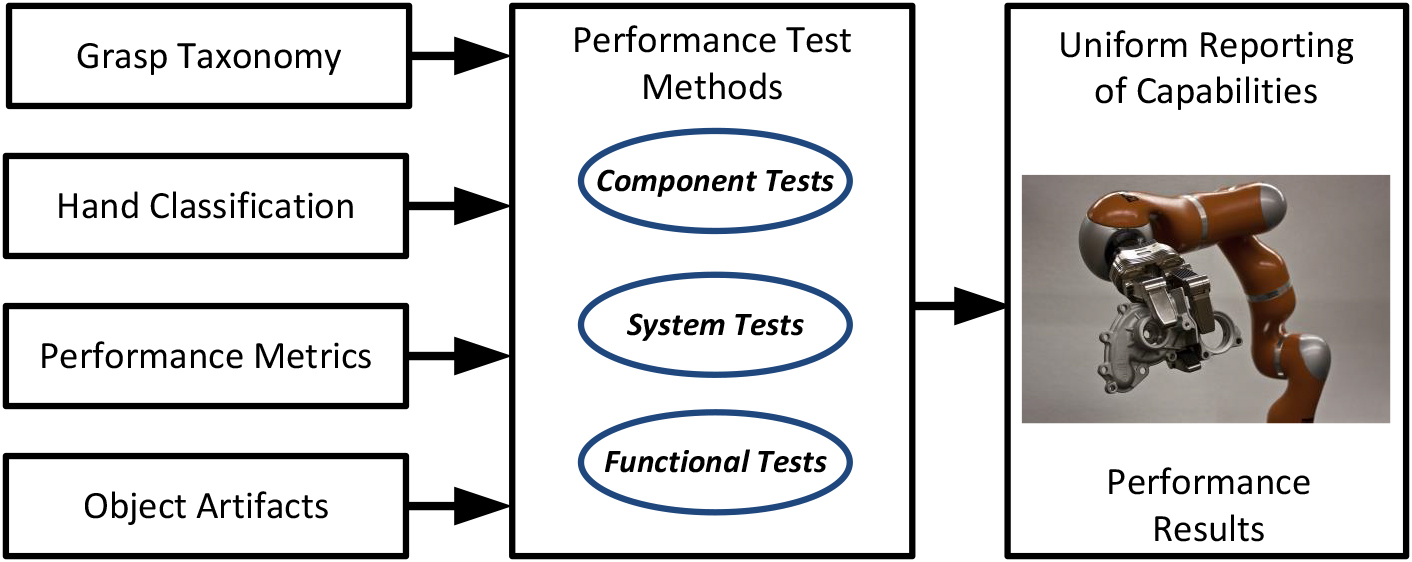}
\caption{Framework for standardized benchmarking of robotic hands}
\label{fig:framework}
\end{figure}

Robotic hands are an integrated, mechatronic system of sensors, motors, and control algorithms typically ranging from three-fingered to five-fingered anthropomorphic designs having both fully and under articulated linkages with compliant or rigid joints. Designs incorporate a variety of sensing technologies, including simple current sensing at the drive motors, load cells, barometers, hydrophones, pressure transducers, electrodes, cameras, and tactile arrays. Depending on the sensory layouts and particular mechanical implementation, tactile sensing capabilities can include the ability to resolve point of contact, directionality and magnitude of contact forces, as well as other sensing modalities such as vibration and temperature. Control algorithms use these signals to incorporate position, velocity, and force control schemes. This wide scope of performance characteristics requires a modular set of performance metrics and associated test methods that can be chosen based on a defined set of grasp types the hand can perform, as well as a scheme for classifying a hand that includes sensing and control capabilities.  Also needed are a common set of test objects (artifacts) to be used along with the test methods. A framework for benchmarking the performance of robotic hands is shown in Figure \ref{fig:framework}.

Grasp taxonomies for the human hand have been developed towards the understanding of grasps humans commonly use in everyday tasks. Cutkosky performed a study of the grasps used by machinists in a small batch manufacturing operation and developed a taxonomy of grasps to provide insights for the design of versatile robotic hands for manufacturing \cite{Cutkosky1989}. Feix et al. derived a taxonomy of grasps based on a literature review of 14 human grasp studies (including Cutkosky's) from both the robotics and medical communities \cite{Feix2009}. The knowledge of these grasp taxonomies has been applied to the design of robotic and prosthetic hands and provides a basis for describing the grasp types a hand can perform.

Performance tests should encompass general grasping tasks as the one defined in 5.2 above and should contain unit, integrated, and functional test methods. When evaluating the capabilities of a robotic hand, unit and integrated performance tests should be agnostic to the other system components such as the robot arm and perception system. While it is possible to access data directly from a robotic hand and derive the defined metrics, these measurements would be based on the inherent properties of the system under test. Therefore, independent measurement systems must be developed to support testing to allow for comparative metrics between systems without effects such as force accuracies and data latencies.

Unit performance characteristics include kinematic properties such as volumetric capabilities and grasp configurations with associated maximum force capabilities. At the very basic level, primitive geometries such as spheres, cylinders, and cubes can be used to characterize the volumetric capabilities of a hand and maximum pinch and grasp forces can be determined at the bounds of these primitive volumetric capabilities. Individual finger tests can be performed to determine the positional accuracy and repeatability of the finger as well as velocity and acceleration characteristics. Sensors can be tested at their stock sensing modalities for properties such as resolution and sensitivity. For example, in the case of tactile sensors, desired characteristics might include normal and shear sensing capabilities as well as the ability to resolve the direction for forces and spatial resolution.

Integrated system characteristics include tests to evaluate the ability of a hand to withstand external forces while maintaining a good level of grasp efficiency and the ability to make initial contact with an object with minimal disturbance. In addition, tests are needed to characterize the integration of a sensor system. For example, tests are needed to characterize feedback latecy from a tactile sensor, as well as to quantify the grasp efficiency of a hand holding onto an object that is subjected to external disturbances.

Functional tests which include a robot arm and perception system can be standardized if they are defined generically to meet the requirements of an application space or to evaluate the capabilities of different robotic hand technologies for a known application. These functional tests would be best suited for a robotic hand that has evolved through a development process that included both unit and integrated testing to the point that warrants testing for use within an intended application space. Finally, functional tests should include testing within the actual application space to determine if a system is capable for the intended application.

In summary, standardized performance benchmarks for robotic hands offer the benefits of an ``honest broker'' to measure how well a system performs in a particular ability where the results of such evaluations and benchmarks help to match capabilities to end-user needs, as well as to help developers improve their product designs. To date, benchmarks to assess the results of grasp research are primarily qualitative measures; however, there is evidence of quantitative assessments of grasping research results scattered across the community. Standardized benchmarks will require a framework for matching the grasp types that a system under test can perform, as well as matching its sensing and control capabilities to the right set of unit and integrated performance tests in order to perform a thorough evaluation of a robotic hand system. General functional tests designed for particular application spaces are also needed for use prior to functional tests to support the actual application.

\newpage

\section*{Symbols used}
\begin{table}[htb]
\centering
\begin{tabular}{|c|l|}
\hline
Symbol & Meaning\\
\hline
$O_{XYZ}$ & inertial reference frame\\
$B_{xyz}$ & body frame attached to the robot\\
$\mathcal{B}$ & object being manipulated\\
$\mathcal{H}$ & robotic hand\\
${\bf u}$ & configuration of a frame in operational space\\
$h$ & number of degrees of freedom of the hand\\
$a$ & number of degrees of freedom of the arm\\ 
$m$ & number of degrees of freedom of the robot ($m=a+h$)\\
${\bf q}$ & configuration of the robot (vector in $\mathbb{R}^m$)\\
${\bf q}_a$ & configuration of the arm (vector in $\mathbb{R}^a$)\\
${\bf q}_h$ & configuration of the hand (vector in $\mathbb{R}^h$)\\
${\bf c}_i$ & contact point between $\mathcal{H}$ and $\mathcal{B}$\\
${\bf w}$ & wrench \\
$\mu, \gamma$ & friction coefficients\\
$f_n$ & normal force exerted at a contact\\
FC & friction cone\\
$n_c$ & number of contact points\\
$\bf J$ & hand Jacobian\\
FK & forward kinematics map\\
${\bf G}$ & grasp matrix\\
$\mathcal{G}$ & multifingered grasp\\
$n$ & number of independent forces and moments\\
{\bf l} & vector of independent forces and moments \\
$\psi$ & gap function\\
\hline
\end{tabular}
\caption{Symbols used in this primer}
\end{table}

\end{document}